\documentclass[runningheads]{llncs}
\usepackage{comment}

\usepackage{epsfig}
\usepackage{subfig}
\usepackage{graphicx}
\usepackage{amsmath}
\usepackage{amssymb}
\usepackage{makecell}
\usepackage{tabu}
\usepackage[flushleft]{threeparttable}
\usepackage{placeins}
\usepackage{longtable}

\usepackage{amsfonts} 
\usepackage{ulem}

\global\long\def\pol{s} 
\global\long\def\cth{\text{CT}} 
\global\long\def\ctp{\text{C}_\text{p}} 
\global\long\def\ctn{\text{C}_\text{n}} 
\global\long\def\eppps{\frac{\text{events}}{\text{pix}\cdot\SI{}{\second}}} 

\newcommand{\fwl}{\text{FWL}}

\newcommand{\eg}[1]{{e.g. #1}}
\newcommand{\ie}[1]{{i.e. #1}}

\newcommand{\etal}{\textit{et al. }}
\newcommand{\dsfont}[1]{\texttt{#1}}
\newcommand{\tablab}[1]{Table~#1}
\newcommand{\figlab}[1]{Figure~#1}
\newcommand{\seclab}[1]{Section~#1}
\newcommand{\eqlab}[1]{eq.~#1}

\usepackage{bm}
\usepackage{booktabs}
\usepackage{makecell}

\usepackage{xcolor}

\usepackage{multirow}
\usepackage{etoolbox}
\usepackage{siunitx}

\robustify\bfseries
\robustify\itshape

\usepackage{cite}
\DeclareSIUnit\revpermin{rpm}

\usepackage[breaklinks=true,bookmarks=false]{hyperref}
\hypersetup{
	colorlinks,
	linkcolor={red!80!black},
	citecolor={green!50!black},
	urlcolor={blue!80!black},
	pdftitle={Reducing the Sim-to-Real Gap for Event Cameras},
	pdfsubject={Computer Vision, Event-based Vision},
	pdfauthor={Timo Stoffregen, Cedric Scheerlinck, Davide Scaramuzza, Tom Drummond, Nick Barnes, Lindsay Kleeman, Robert Mahony},
	pdfkeywords={Event cameras, Dynamic Vision Sensor, Image reconstruction, Optic flow, Learning, Neural Network}
}

\usepackage[breaklinks=true,bookmarks=false]{hyperref}
\usepackage[printwatermark]{xwatermark}
\newwatermark[allpages,color=gray!50,scale=0.4,xpos=0,ypos=120]{This paper has been accepted for publication at the European Conference on Computer Vision, 2020}
\begin{document}

\pagestyle{headings}
\mainmatter
\def\ECCVSubNumber{5855}  

\title{Reducing the Sim-to-Real Gap for Event Cameras} 

\titlerunning{Reducing the Sim-to-Real Gap for Event Cameras}

\author{
	Timo Stoffregen$^{\star,}$\inst{1,3} \and
	Cedric Scheerlinck\thanks{Equal contribution.}$^,$\inst{2,3} \and
	Davide Scaramuzza \inst{4} \and
	Tom Drummond \inst{1,3} \and
	Nick Barnes \inst{2,3} \and
	Lindsay Kleeman \inst{1} \and
	Robert Mahony \inst{2,3}}
\authorrunning{T. Stoffregen, C. Scheerlinck et al.}

\institute{
	Dept. of ECSE,
	Monash University, Melbourne, Australia \and
	Australian National University, Canberra, Australia \and
	Australian Centre for Robotic Vision, Australia \and
	University of Zurich, Zurich, Switzerland}

\maketitle

\begin{abstract}

Event cameras are paradigm-shifting novel sensors that report asynchronous, per-pixel brightness changes called `events' with unparalleled low latency.
This makes them ideal for high speed, high dynamic range scenes where conventional cameras would fail. 
Recent work has demonstrated impressive results using Convolutional Neural Networks (CNNs) for video reconstruction and optic flow with events.
We present strategies for improving training data for event based CNNs that result in 20-\SI{40}{\percent} boost in performance of existing state-of-the-art (SOTA) video reconstruction networks retrained with our method, and up to \SI{15}{\percent} for optic flow networks.
A challenge in evaluating event based video reconstruction is lack of quality ground truth images in existing datasets.
To address this, we present a new \textbf{High Quality Frames (HQF)} dataset, containing events and ground truth frames from a DAVIS240C that are well-exposed and minimally motion-blurred.
We evaluate our method on HQF + several existing major event camera datasets.
\end{abstract}

\begin{center}
\textbf{Video, code and datasets: \url{https://timostoff.github.io/20ecnn}}
\end{center}
\begin{figure}
	\newcommand{\widthplot}{0.18\textwidth}
	\centering
	\renewcommand*{\arraystretch}{1.1}
	\setlength{\tabcolsep}{0.3ex} %
\begin{tabular}{lccccc}
	\rotatebox{90}{Groundtruth} & \includegraphics[width=\widthplot]{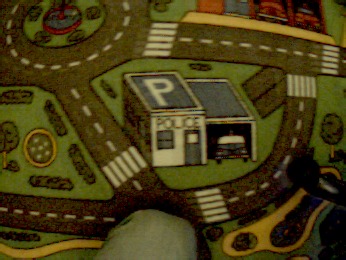} &
	\includegraphics[width=\widthplot]{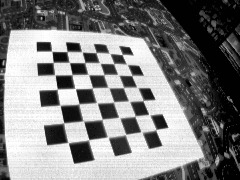} &
	\includegraphics[width=\widthplot]{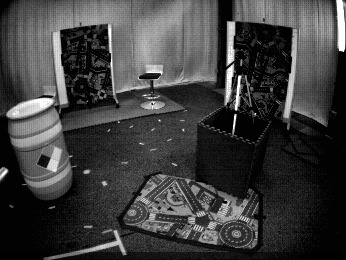} &
	\includegraphics[width=\widthplot]{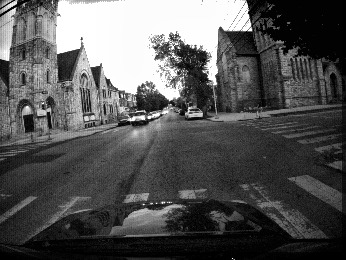} &
	\includegraphics[width=\widthplot]{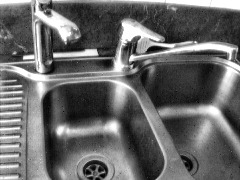} \\
	
	\rotatebox{90}{SOTA} &
	\includegraphics[width=\widthplot]{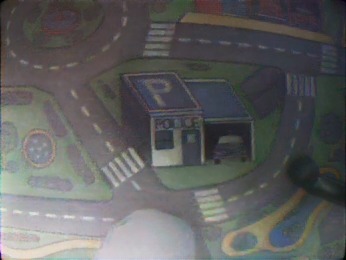} &
	\includegraphics[width=\widthplot]{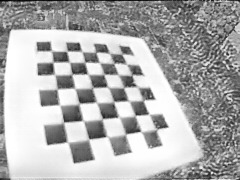} &
	\includegraphics[width=\widthplot]{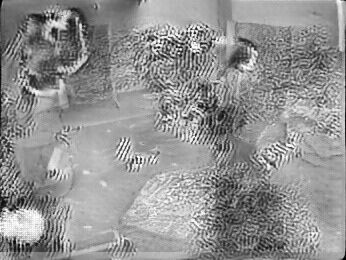} &
	\includegraphics[width=\widthplot]{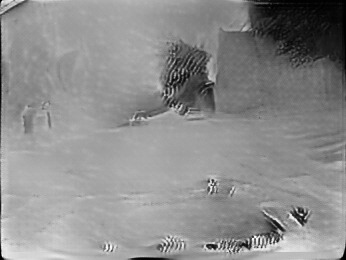} &
	\includegraphics[width=\widthplot]{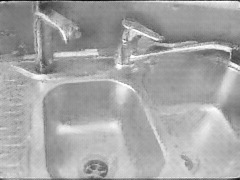} \\
	
	\rotatebox{90}{\textbf{Ours}} &
	\includegraphics[width=\widthplot]{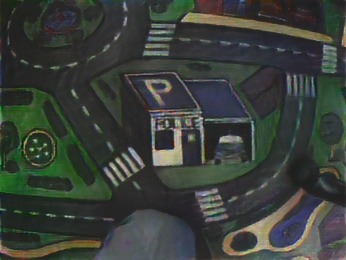} &
	\includegraphics[width=\widthplot]{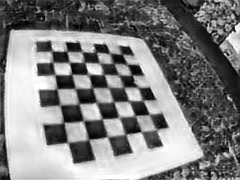} &
	\includegraphics[width=\widthplot]{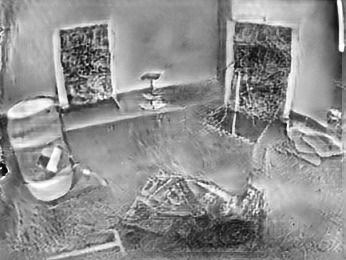} &
	\includegraphics[width=\widthplot]{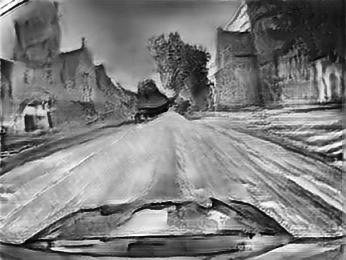} &
	\includegraphics[width=\widthplot]{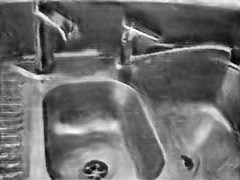} \\
	
	& CED & IJRR & MVSEC & MVSEC & \textbf{HQF}
\end{tabular}

	\caption{Top: ground truth reference image.
		Middle/bottom: state-of-the-art E2VID \cite{Rebecq20pami} vs. our reconstructed images from events only. Challenging scenes from event camera datasets: CED \cite{Scheerlinck19cvprw}, IJRR \cite{Mueggler17ijrr}, MVSEC \cite{Zhu18ral} and our \textbf{HQF} dataset.
	}
	\label{fig:front_page}
\end{figure}

\section{Introduction}
Event-based cameras such as the Dynamic Vision Sensor (DVS) \cite{Lichtsteiner08ssc} are novel, bio-inspired visual sensors.
Presenting a paradigm-shift in visual data acquisition, pixels in an event camera operate by asynchronously and independently reporting intensity changes in the form of events, represented as a tuple of $x, y$ location, timestamp $t$ and polarity of the intensity change $\pol$.
By moving away from fixed frame-rate sampling of conventional cameras, event cameras deliver several key advantages in terms of low power usage (in the region of \SI{5}{\milli\watt}), high dynamic range (\SI{140}{\decibel}), low latency and timestamps with resolution on the order of \SI{}{\micro\second}.

With the recent preponderance of deep learning techniques in computer vision, the question of how to apply this technology to event data has been the subject of several recent works.
Zhu \etal \cite{Zhu18rss} propose an unsupervised network able to learn optic flow from real event data, while Rebecq \etal \cite{Rebecq19cvpr, Rebecq20pami} showed that supervised networks trained on synthetic events transferred well to real event data.
Simulation shows promise since data acquisition and ground truth are easily obtainable, in contrast to using real data.
However, mismatch between synthetic and real data degrades performance, so a key challenge is simulating realistic data.

We generate training data that better matches real event camera data by analyzing the statistics of existing datasets to inform our choice of simulation parameters.
A major finding is that the contrast threshold ($\cth$) - the minimum change in brightness required to trigger an event - is a key simulation parameter that impacts performance of supervised CNNs.
Further, we observe that the apparent contrast threshold of real event cameras varies greatly, even within one dataset.
Previous works such as event based video reconstruction \cite{Rebecq20pami} choose contrast thresholds that work well for some datasets, but fail on others.
Unsupervised networks trained on real data such as event based optic flow \cite{Zhu18rss} may be retrained to match any real event camera - at the cost of new data collection and training.
We show that using $\cth$ values for synthetic training data that are correctly matched to $\cth$s of real datasets is a key driver in improving performance of retrained event based video reconstruction and optic flow networks across multiple datasets.
We also propose a simple noise model which yields up to \SI{10}{\percent} improvement when added during training.

A challenge in evaluating image and video reconstruction from events is lack of quality ground truth images registered and time-synchronized to events, because most existing datasets focus on scenarios where event cameras excel (high speed, HDR) and conventional cameras fail.
To address this limitation, we introduce a new High Quality Frames (HQF) dataset that provides several sequences in well lit environments with minimal motion blur.
These sequences are recorded with a DAVIS240C event camera that provides perfectly aligned frames from an integrated Active Pixel Sensor (APS).
HQF also contains a diverse range of motions and scene types, including slow motion and pauses that are challenging for event based video reconstruction.
We quantitatively evaluate our method on two major event camera datasets: IJRR \cite{Mueggler17ijrr} and MVSEC \cite{Zhu18ral}, in addition to our HQF, demonstrating gains of \SI{20}{}-\SI{40}{\percent} for video reconstruction and up to \SI{15}{\percent} for optic flow when we retrain existing SOTA networks.

\textit{Contribution}
We present a method to generate synthetic training data that improves generalizability to real event data, guided by statistical analysis of existing datasets.
We additionally propose a simple method for dynamic train-time noise augmentation that yields up to \SI{10}{\percent} improvement for video reconstruction.
Using our method, we retrain several network architectures from previously published works on video reconstruction \cite{Rebecq20pami,Scheerlinck20wacv} and optic flow \cite{Zhu18rss,Zhu19cvpr} from events.
We are able to show significant improvements that persist over architectures and tasks.
Thus, we believe our findings will provide invaluable insight for others who wish to train models on synthetic events for a variety of tasks.
We provide a new comprehensive High Quality Frames dataset targeting ground truth image frames for video reconstruction evaluation.
Finally, we provide our data generation code, training set, training code and our pretrained models, together with dozens of useful helper scripts for the analysis of event-based datasets to make this task easier for fellow researchers.

In summary, our major contributions are:
\begin{itemize}
	\item A method for simulating training data that yields \SI{20}{\percent}-\SI{40}{} and up to \SI{15}{\percent} improvement for event based video reconstruction and optic flow CNNs.
	\item Dynamic train-time event noise augmentation.
	\item A novel High Quality Frames dataset.
	\item Extensive analysis and evaluation of our method.
    \item An optic flow evaluation metric \textit{Flow Warp Loss ($\fwl$)}, tailored to event data, that does not require ground truth flow.
	\item Open-source code, training data and pretrained models.
\end{itemize}

The remainder of the paper is as follows.
\seclab{\ref{sec:related_works}} reviews related works.
\seclab{\ref{sec:method}} outlines our method for generating training data, training and evaluation, and introduces our HQF dataset.
\seclab{\ref{sec:experiments}} presents experimental results on video reconstruction and optic flow.
\seclab{\ref{sec:discussion}} discusses our major findings and concludes the paper.

\section{Related Works}
\label{sec:related_works}

\subsection{Video Reconstruction}
Video and image reconstruction from events has been a popular topic in the event based vision literature.
Several approaches have been proposed in recent years; Kim \etal \cite{Kim14bmvc} used an EKF to reconstruct images from a rotating event camera, later extending this approach to full 6-DOF camera motions \cite{Kim16eccv}.
Bardow \etal \cite{Bardow16cvpr} used a sliding spatiotemporal window of events to simultaneously optimize both optic flow and intensity estimates using the primal-dual algorithm, although this method remains sensitive to hyperparameters.
Reinbacher \etal \cite{Reinbacher16bmvc} proposed direct integration with periodic manifold regularization on the Surface of Active Events (SAE \cite{Mueggler15icra}) to reconstruct video from events.
Scheerlinck \etal \cite{Scheerlinck18accv, Scheerlinck19ral} achieved computationally efficient, continuous-time video reconstruction via complementary and high-pass filtering.
This approach can be combined with conventional frames, if available, to provide low frequency components of the image.
However, if taken alone, this approach suffers from artifacts such as ghosting effects and bleeding edges.

Recently, convolutional neural networks (CNNs) have been brought to bear on the task of video reconstruction.
Rebecq \etal \cite{Rebecq19cvpr,Rebecq20pami} presented E2VID, a recurrent network that converts events (discretized into a voxel grid) to video.
A temporal consistency loss based on \cite{Lai18eccv} was introduced to reduce flickering artifacts in the video, due to small differences in the reconstruction of subsequent frames.
E2VID is current state-of-the-art.
Scheerlinck \etal were able to reduce model complexity by \SI{99}{\percent} with the \textit{FireNet} architecture \cite{Scheerlinck20wacv}, with only minor trade-offs in reconstruction quality, enabling high frequency inference.

\subsection{Optic Flow}
Since event based cameras are considered a good fit for applications involving motion \cite{Gallego19arxiv}, much work has been done on estimating optic flow with event cameras  \cite{Benosman12nn, Benosman14tnnls, Brosch15fns, Bardow16cvpr, Gallego18cvpr, Stoffregen17acra, Stoffregen19iccv, Liu18bmvc, Almatrafi19tci}. 
Recently, Zhu \etal proposed a CNN (EV-FlowNet) for estimating optic flow from events \cite{Zhu18rss}, together with the Multi-Vehicle Stereo Event Camera (MVSEC) dataset \cite{Zhu18ral} that contains ground truth optic flow estimated from depth and ego-motion sensors.
The input to EV-FlowNet is a 4-channel image formed by recording event counts and the most recent timestamps for negative and positive events.
The loss imposed on EV-FlowNet was an image-warping loss \cite{Yu16eccvw} that took photometric error between subsequent APS frames registered using the predicted flow.
A similar approach was taken by Ye \etal \cite{Ye19arxiv}, in a network that estimated depth and camera pose to calculate optic flow.
In \cite{Zhu19cvpr}, Zhu \etal improved on prior work by replacing the image-warping loss with an event-warping loss that directly transports events to a reference time using the predicted flow.
We use a similar method to evaluate optic flow performance of several networks (see \seclab{\ref{ssec:flow}}).
Zhu \etal \cite{Zhu19cvpr} also introduced a novel input representation based on event discretization that places events into bins with temporal bilinear interpolation to produce a voxel grid.
EV-FlowNet was trained on data from MVSEC \cite{Zhu18ral} and Ye \etal \cite{Ye19arxiv} even trained, then validated on the same sequences; our results (\seclab{\ref{sec:evaluation}}) indicate that these networks suffer from overfitting.

\subsection{Input Representations}
To use conventional CNNs, events must first be transformed into an amenable grid-based representation.
While asynchronous spiking neural networks can process raw events and have been used for object recognition \cite{Lee16fns, Orchard15pami, PerezCarrasco13pami} and optic flow \cite{Benosman12nn, Benosman14tnnls}, lack of appropriate hardware or effective error backpropagation techniques renders them yet uncompetitive with state-of-the-art CNNs.
Several grid-based input representations for CNNs have been proposed: simple event images \cite{Maqueda18cvpr, Zhu18rss} (events are accumulated to form an image), Surface of Active Events (SAE) \cite{Zhu18rss} (latest timestamp recorded at each pixel), Histogram of Averaged Time Surfaces (HATS) \cite{Sironi18cvpr} and even learned input representations, where events are sampled into a grid using convolutional kernels \cite{Gehrig19iccv}.
Zhu \etal \cite{Zhu19cvpr} and Rebecq \etal \cite{Rebecq20pami} found best results using a voxel grid representation of events, where the temporal dimension is essentially discretized and subsequently binned into an $n$ dimensional grid (\eqlab{\ref{eq:voxel}}).

\section{Method}
\label{sec:method}

\subsection{Event Camera Contrast Threshold}
\label{sec:choice_of_ct}
In an ideal event camera, a pixel at $(x, y)$ triggers an event $e_i$ at time $t_i$ when the brightness since the last event $e_{i-1}$ at that pixel changes by a threshold $C$, given $t-t_{i-1}>r$, the refractory period of that pixel.
$C$ is referred to as the contrast threshold ($\cth$) and can be typically adjusted in modern event cameras.
In reality, the values for $C$ are not constant in time nor homogeneous over the image plane nor is the positive threshold $C_p$ necessarily equal to the negative threshold $C_n$.
In simulation (\eg using ESIM \cite{Rebecq18corl}), $\cth$s are typically sampled from $\mathcal{N}(\mu$=0.18, $\sigma$=0.03$)$ to model this variation \cite{Rebecq19cvpr, Rebecq20pami, Gehrig19iccv}.
The $\cth$ is an important simulator parameter since it determines the number and distribution of events generated from a given scene.
\global\long\def\ct_dataset_stats_width{0.33\linewidth}
\begin{figure}[t]
\centering
\resizebox{\textwidth}{!}{%
\subfloat[IJRR/MVSEC vs. ESIM\label{fig:ct_dataset_stats_a}]{{\includegraphics[width=\ct_dataset_stats_width]{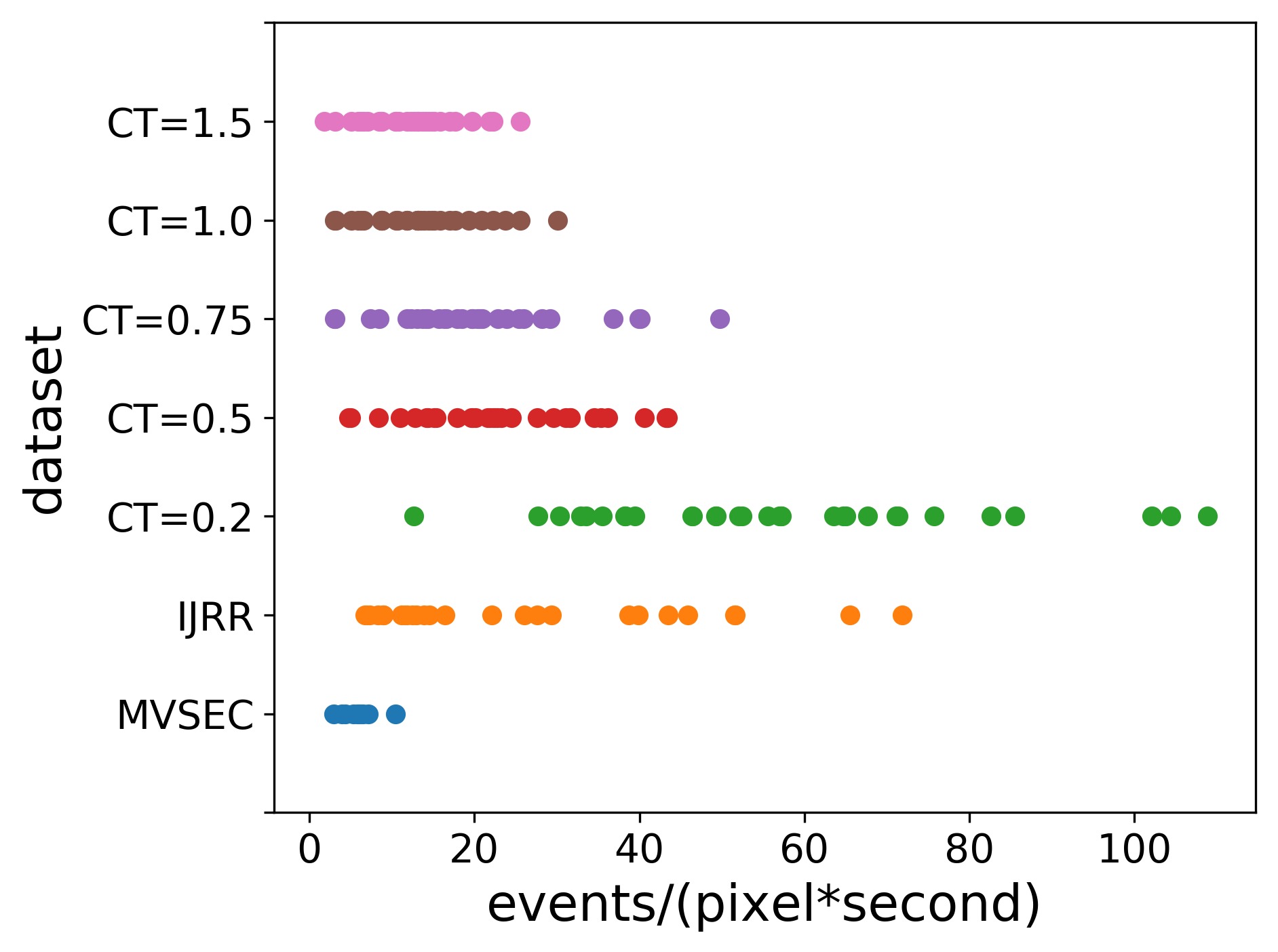}}}~~~
\subfloat[IJRR vs ESIM\label{fig:ct_dataset_stats_b}]{{\includegraphics[width=\ct_dataset_stats_width]{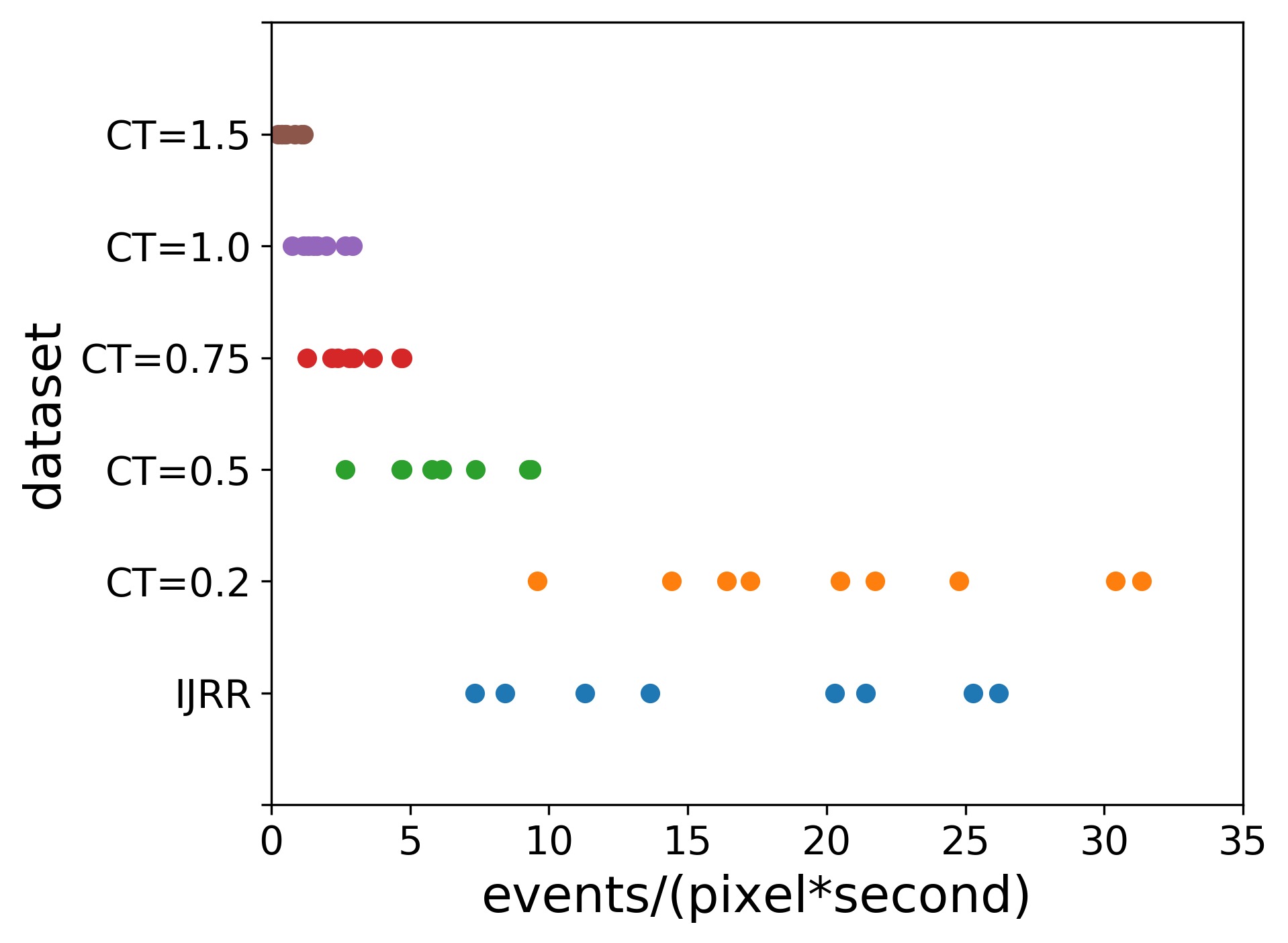}}}~~~
\subfloat[MVSEC vs ESIM\label{fig:ct_dataset_stats_c}]{{\includegraphics[width=\ct_dataset_stats_width]{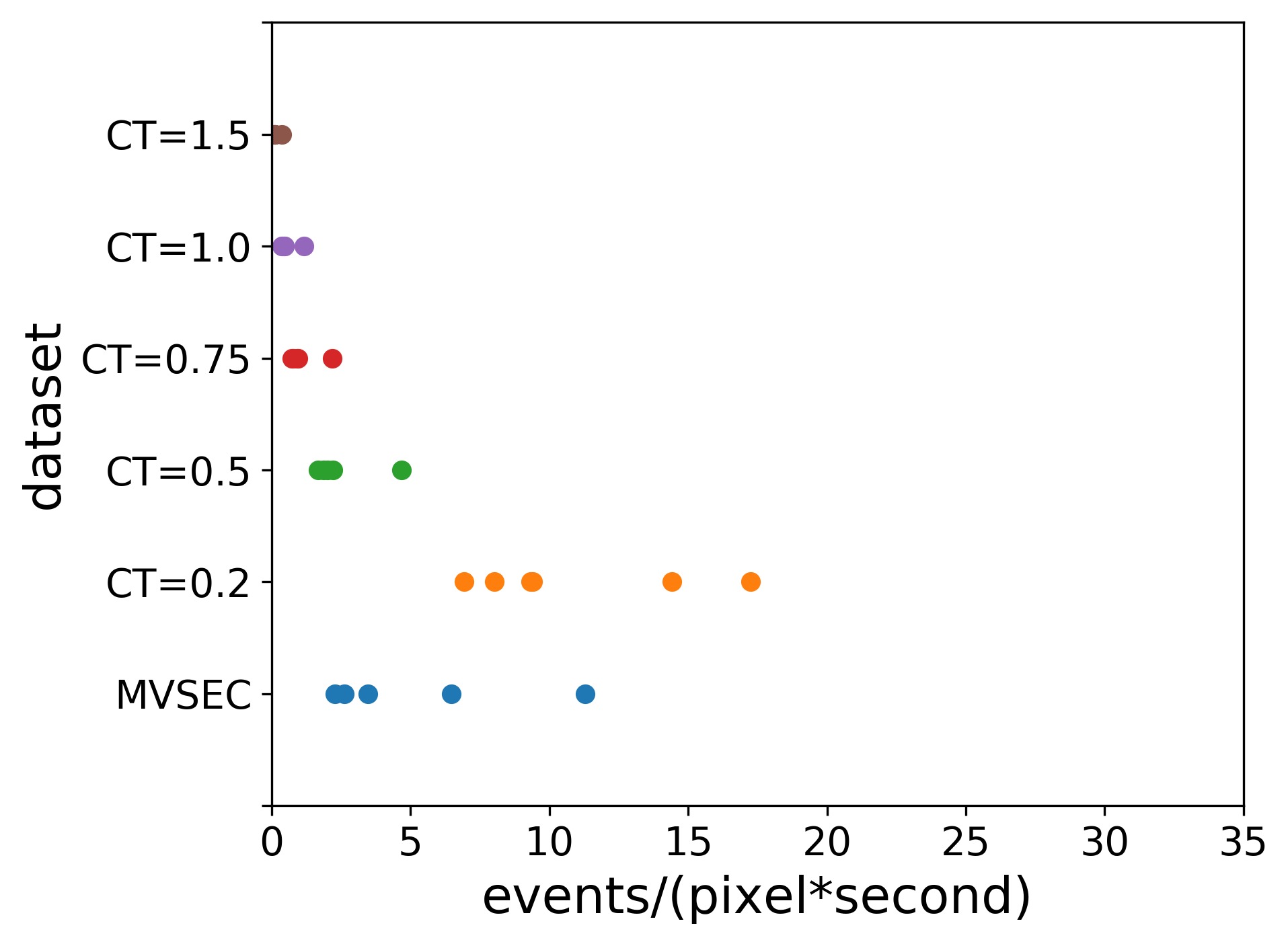}}}
}
\caption{
   	\label{fig:ct_dataset_stats}
    Each dot represents a sequence from the given dataset (y-axis).
	(a) $\eppps$ of IJRR and MVSEC vs. ESIM training datasets ($\cth$ 0.2-1.5) described in Section \ref{sec:training_data}. 
    (b)
	$\eppps$ of IJRR vs. ESIM events simulated from IJRR \textit{APS frames}.
	(c) $\eppps$ of MVSEC vs. ESIM events simulated from MVSEC \textit{APS frames}.
    }
\end{figure}

While the real $\cth$s of previously published datasets are unknown, one method to estimate $\cth$s is via the proxy measurement of average events per pixel per second ($\eppps$).
Intuitively, higher $\cth$s tend to reduce the $\eppps$ for a given scene.
While other methods of $\cth$ estimation exist (see supp. material), we found that tuning the simulator $\cth$s to match $\eppps$ of real data worked well.
Since this measure is affected by scene dynamics (\ie faster motions increase $\eppps$ independently of $\cth$), we generated a diverse variety of realistic scene dynamics.
The result of this experiment (\figlab{\ref{fig:ct_dataset_stats_a}}) indicates that a contrast threshold setting of between $0.2$ and $0.5$ would be more appropriate for sequences from the IJRR dataset \cite{Mueggler17ijrr}.
The larger diversity of motions is also apparent in the large spread of the $\eppps$ compared to MVSEC \cite{Zhu18ral} whose sequences are tightly clustered.

\begin{figure}[t]
\global\long\def\dataset_quality_comparison_width{0.19\linewidth}
\centering
\begin{tabular}{c c c c c}
	\frame{\includegraphics[width=\dataset_quality_comparison_width]{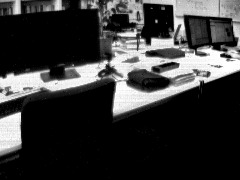}} &
	\frame{\includegraphics[width=\dataset_quality_comparison_width]{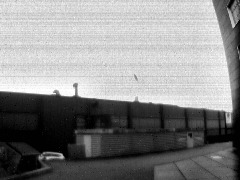}} &
	\frame{\includegraphics[width=\dataset_quality_comparison_width]{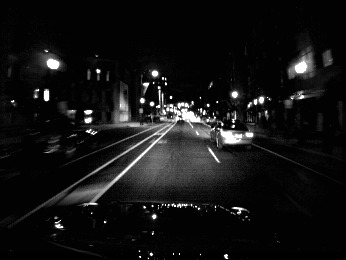}} &
	\frame{\includegraphics[width=\dataset_quality_comparison_width]{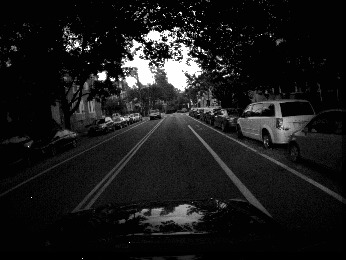}} &
	\frame{\includegraphics[width=\dataset_quality_comparison_width]{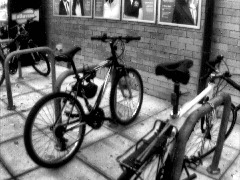}}
	\\
	\frame{\includegraphics[width=\dataset_quality_comparison_width]{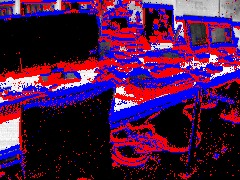}} & \frame{\includegraphics[width=\dataset_quality_comparison_width]{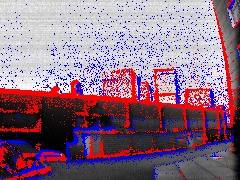}} &
	\frame{\includegraphics[width=\dataset_quality_comparison_width]{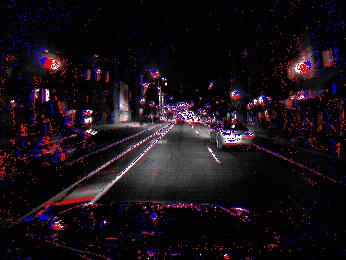}} &
	\frame{\includegraphics[width=\dataset_quality_comparison_width]{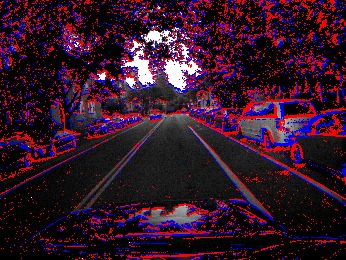}} &
	\frame{\includegraphics[width=\dataset_quality_comparison_width]{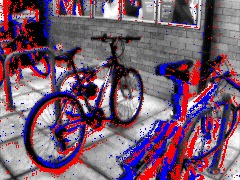}}
	\\
	(a) IJRR & (b) IJRR & (c) MVSEC & (d) MVSEC & (e) HQF
\end{tabular}
\caption{\label{fig:dataset_quality_comparison} 
	Note that in many sequences from the commonly used IJRR and MVSEC datasets, the accompanying APS frames are of low quality.
	The top row shows the APS frames, the bottom row overlays the events.
	As can be seen, many features are not visible in the APS frames, making quantitative evaluation difficult.
	This motivates our own High Quality Frames dataset (HQF).
}
\end{figure}

As an alternative experiment to determine $\cth$s of existing datasets, we measured the $\eppps$ of events simulated using the actual APS (ground truth) frames of IJRR and MVSEC sequences.
Given high quality images with minimal motion blur and little displacement, events can be simulated through image interpolation and subtraction.
Given an ideal image sequence, the simulator settings should be tunable to get the exact same $\eppps$ from simulation as from the real sensor.
Unfortunately APS frames are not usually of a very high quality (\figlab \ref{fig:dataset_quality_comparison}), so we were limited to using this approach on carefully curated snippets (\figlab{\ref{fig:exposure_ijrr_mvsec}}).
The results of this experiment in \figlab{\ref{fig:ct_dataset_stats_b} } and \ref{fig:ct_dataset_stats_c} indicate similar results of lower contrast thresholds for IJRR and higher for MVSEC, although accuracy is limited by the poor quality APS frames.

\global\long\def\exposure_ijrr_mvsec_width{0.495\linewidth}
\begin{figure}[t]
\centering
\resizebox{\textwidth}{!}{%
\subfloat[Poorly exposed from IJRR and MVSEC]{\frame{\label{fig:exposure_ijrr_mvsec:a}\includegraphics[width=\exposure_ijrr_mvsec_width]{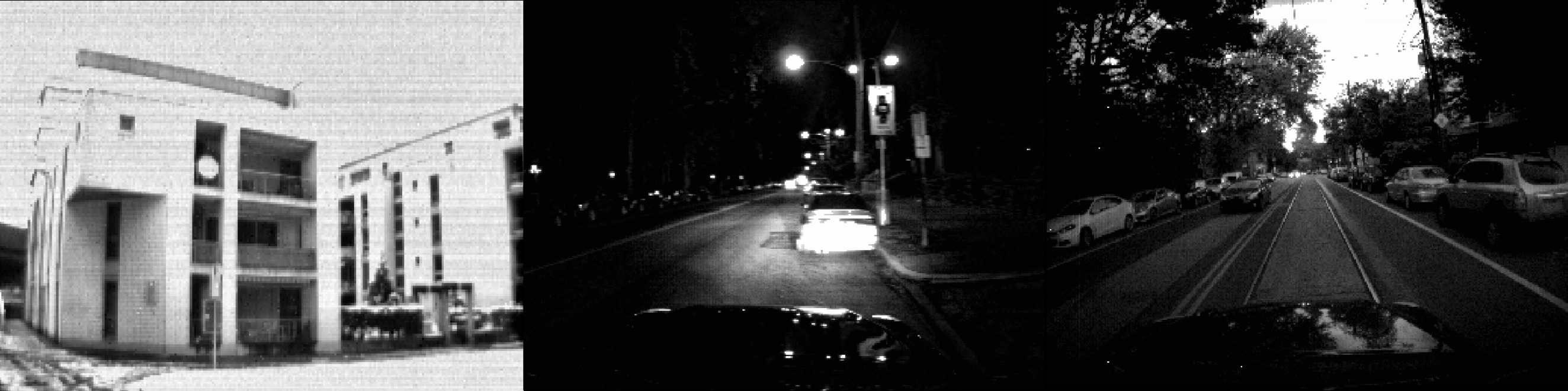}}}~
\subfloat[Well exposed from IJRR and MVSEC]{\frame{\label{fig:exposure_ijrr_mvsec:b}\includegraphics[width=\exposure_ijrr_mvsec_width]{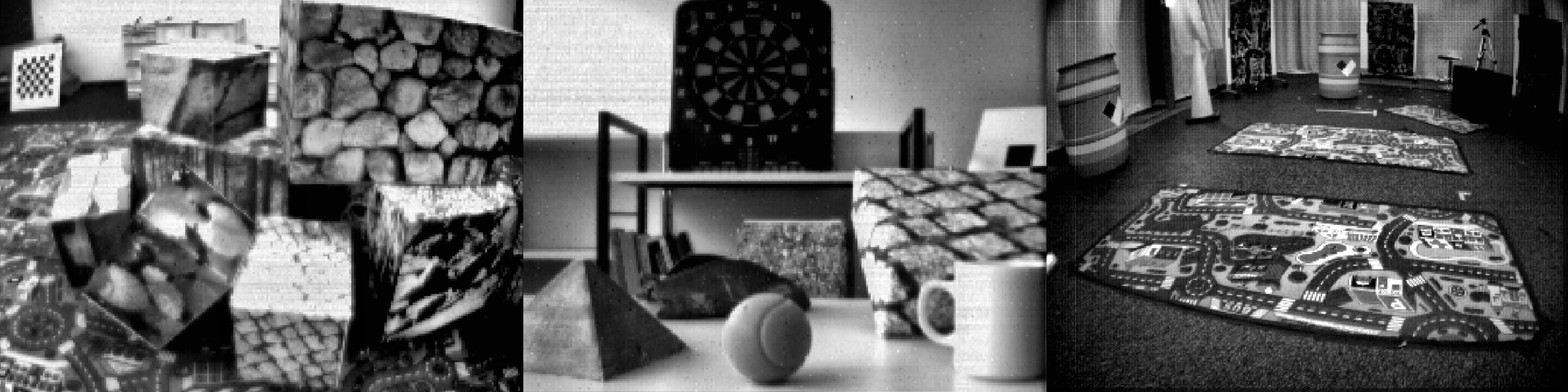}}}
}
    \caption{\label{fig:exposure_ijrr_mvsec} 
    Examples of frames from IJRR and MVSEC after local histogram equalization, with poorly exposed sequences in \ref{fig:exposure_ijrr_mvsec:a}, and better exposed images in \ref{fig:exposure_ijrr_mvsec:b}.
    }
\end{figure}

\subsection{Training Data}
\label{sec:training_data}
We used an event camera simulator, ESIM \cite{Rebecq18corl} to generate training sequences for our network.
There are several modes of simulation available, of which we used ``Multi-Object-2D'' that facilitates moving images in simple 2D motions, restricted to translations, rotations and dilations over a planar background.
This generates sequences reminiscent of Flying Chairs \cite{Dosovitskiy15iccv}, where objects move across the screen at varying velocities.
In our generation scheme, we randomly selected images from COCO \cite{TsungYi14eccv} and gave them random trajectories over the image plane.
Our dataset contains 280 sequences, \SI{10}{\second} in length.
Sequences alternate between four archetypal scenes; slow motion with 0-5 foreground objects, medium speed motion with 5-10 foreground objects, fast speed with 5-20 foreground objects and finally, full variety of motions with 10-30 foreground objects.
This variety encourages networks to generalize to arbitrary real world camera motions, since a wide range of scene dynamics are presented during training.
Sequences were generated with contrast thresholds ($\cth$s) between 0.1 and 1.5 in ascending order.
Since real event cameras do not usually have perfectly balanced positive and negative thresholds, the positive threshold $\ctp=\ctn\cdot x, x\in\mathcal{N}(\mu$=1.0, $\sigma$=0.1$)$.

The events thus generated were discretized into a voxel grid representation.
In order to ensure synchronicity with the ground truth frames of our training set and later with the ground truth frames of our validation set, we always take all events between two frames to generate a voxel grid.
Given $N$ events $e_i=\{x_i,y_i,t_i,\pol_i \}_{i=0,...,N}$ spanning $\Delta_T=t_N-t_0$ seconds, a voxel grid $V$ with $B$ bins can be formed through temporal bilinear interpolation via
\begin{equation}
\label{eq:voxel}
V_{k\in[0,B-1]}=\sum_{i=0}^N \pol_i \max (0, 1-|t_i^*-k|)
\end{equation}
where $t_i^*$ is the timestamp normalized to the range $[0, B-1]$ via $t_i^*=\frac{t_i-t_0}{\Delta_T}(B-1)$ and the bins are evenly spaced over the range $[t_0, t_N]$.
This method of forming voxels has some limitations; it is easy to see that the density of the voxels can vary greatly, depending on the camera motion and frame rate of the camera.
Thus, it is important to train the network on a large range of event rates $\eppps$ and voxel densities.
During inference, other strategies of voxel generation can be employed, as further discussed in the supplementary materials.
We used $B=5$ throughout the experiments in this paper.
In earlier experiments we found values of $B=2,5,15$ produced no significant differences.

\subsection{Sequence Length}
To train recurrent networks, we sequentially passed $L$ inputs to the network and computed the loss for each output.
Finally, the losses were summed and a backpropagation update was performed based on the gradient of the final loss with respect to the network weights.
Since recurrent units in the network are initialized to zero, lower values of $L$ restrict the temporal support that the recurrent units see at train time.
To investigate the impact of sequence length $L$, we retrain our networks using $L=40$ (as in E2VID \cite{Rebecq20pami}) and $L=120$.
In the case of non-recurrent networks such as EV-FlowNet \cite{Zhu18rss,Zhu19cvpr}, we ignore the sequence length parameter.

\subsection{Loss}
For our primary video reconstruction loss function we used ``learned perceptual image patch similarity'' (LPIPS) \cite{Zhang18cvprLPIPS}.
LPIPS is a fully differentiable similarity metric between two images that compares hidden layer activations of a pretrained network (\eg Alex-Net or VGG), and was shown to better match human judgment of image similarity than photometric error or SSIM \cite{Wang04tip}.
Since our event tensors were synchronized to the ground truth image frames by design (the final event in the tensor matches the frame timestamp), we computed the LPIPS distance between our reconstruction and the corresponding ground truth frame.
As recommended by the authors \cite{Zhang18cvprLPIPS}, we used the Alex-Net variant of LPIPS.
We additionally imposed a temporal consistency loss \cite{Lai18eccv} that measures photometric error between consecutive images after registration based on optic flow, subject to an occlusion mask.
For optic flow, we used the L1 distance between our prediction and ground truth as the training loss.

\subsection{Data Augmentation}
During training, Rebecq \etal \cite{Rebecq20pami} occasionally set the input events to zero and performed a forward-pass step within a sequence, using the previous ground truth image frame to compute the loss.
The probability of initiating a pause when the sequence is running $P(p|r)=0.05$, while the probability of maintaining the paused state when the sequence is already paused $P(p|p)=0.9$ to encourage occasional long pauses.
This encourages the recurrent units of the network to learn to `preserve' the output image in absence of new events.
We used pause augmentation to train all recurrent networks.

Event cameras provide a noisy measurement of brightness change, subject to background noise, refractory period after an event and hot pixels that fire many spurious events.
To simulate real event data, we applied a refractory period of 1ms.
At train time, for each sequence of $L$ input event tensors we optionally added zero-mean Gaussian noise ($\mathcal{N}(\mu$=0, $\sigma$=0.1$)$) to the event tensor to simulate uncorrelated background noise, and randomly elected a few `hot' pixels.
The number of hot pixels was drawn from a uniform distribution from 0 to 0.0001, multiplied by the total number of pixels.
Hot pixels have a random value ($\mathcal{N}(\mu$=0, $\sigma$=0.1$)$) added to every temporal bin in each event tensor within a sequence.
To determine whether augmenting the training data with noise benefits performance on real data, we retrained several models with and without noise (\tablab{\ref{tab:ablation_all}}).

\subsection{Architecture}
To isolate the impact of our method from choice of network architecture, we retrained state-of-the-art (SOTA) video reconstruction network E2VID \cite{Rebecq20pami} and SOTA optic flow network EV-FlowNet described in \cite{Zhu18rss,Zhu19cvpr}.
Thus, differences in performance for each task are not due to architecture.
Additionally, we aim to show that our method generalizes to multiple architectures.
While we believe architecture search may further improve results, it is outside the scope of this paper.

\subsection{High Quality Frames Dataset}
\label{sec:hqf}
To evaluate event camera image reconstruction methods, we compared reconstructed images to temporally synchronized, registered ground truth reference images.
Event cameras such as the DAVIS \cite{Brandli14ssc} can capture image frames (in addition to events) that are timestamped and registered to the events, that may serve as ground truth.
Previous event camera datasets such as IJRR \cite{Mueggler17ijrr} and MVSEC \cite{Zhu18ral} contain limited high quality DAVIS frames, while many frames are motion-blurred and or under/overexposed (\figlab{\ref{fig:dataset_quality_comparison}}).
As a result, Rebecq \etal \cite{Rebecq20pami} manually rejected poor quality frames, evaluating on a smaller subset of IJRR.

We present a new High Quality Frames dataset (HQF) aimed at providing ground truth DAVIS frames that are minimally motion-blurred and well exposed.
In addition, our HQF covers a wider range of motions and scene types than the evaluation dataset used for E2VID, including: static/dynamic camera motion vs. dynamic camera only, very slow to fast vs. medium to fast and indoor/outdoor vs. indoor only.
To record HQF, we used two different DAVIS240C sensors to capture data with different noise/$\cth$ characteristics.
We used default bias settings loaded by the RPG DVS ROS driver%
\footnote{\href{https://github.com/uzh-rpg/rpg_dvs_ros}{https://github.com/uzh-rpg/rpg\_dvs\_ros}},
and set exposure to either auto or fixed to maximize frame quality.
Our HQF provides temporally synchronized, registered events and DAVIS frames (further details in supplementaries, \tablab 6).

\section{Experiments}
\label{sec:experiments}

\subsection{Evaluation}
\label{sec:evaluation}
We evaluated our method by retraining two state-of-the-art event camera neural networks: E2VID \cite{Rebecq19cvpr,Rebecq20pami}, and EV-FlowNet \cite{Zhu18rss,Zhu19cvpr}.
Our method outperforms previous state-of-the-art in image reconstruction and optic flow on several publicly available event camera datasets including IJRR \cite{Mueggler17ijrr} and MVSEC \cite{Zhu18ral}, and our new High Quality Frames dataset (HQF, \seclab{\ref{sec:hqf}}).

For video reconstruction on the datasets HQF, IJRR and MVSEC (\tablab{\ref{tab:eval:all_in_one}}) we obtained a \SI{40}{\percent}, \SI{20}{\percent} and \SI{28}{\percent} improvement over E2VID \cite{Rebecq20pami} respectively, using LPIPS.
For optic flow we obtained a \SI{12.5}{\percent}, \SI{10}{\percent} and \SI{16}{\percent} improvement over EV-FlowNet \cite{Zhu18rss} on flow warp loss ($\fwl$, \eqlab{\ref{eq:fwl}}).
Notably, EV-FlowNet was trained on MVSEC data (\dsfont{outdoor\_day2} sequence), while ours was trained entirely on synthetic data, demonstrating the ability of our method to generalize to real event data.

\begin{table}
		\caption{
		Comparison of state-of-the-art methods of video reconstruction and optic flow to networks trained using our dataset on HQF, IJRR and MVSEC. Best in bold.
	}
	\resizebox{\textwidth}{!}{%
	\begin{threeparttable}
	\newcolumntype{Z}{S[table-format=2.2,table-auto-round]}
	\centering
	\setlength{\tabcolsep}{0.6mm}
	\renewcommand*{\arraystretch}{1.1}
	\newcommand{\tinyspace}{1mm}
	\newcommand{\smallspace}{2mm}
	\small
	\begin{tabular}{@{}lcZZcZZcZZcZZcr@{}}
		\toprule
		\multirow{2}[3]{*}{Sequence} && \multicolumn{2}{c}{MSE} && \multicolumn{2}{c}{SSIM} && \multicolumn{2}{c}{LPIPS} && \multicolumn{2}{c}{FWL} \\
		\cmidrule(l{\tinyspace}r{\smallspace}){3-4} 		\cmidrule(l{\tinyspace}r{\smallspace}){6-7} 
		\cmidrule(l{\tinyspace}r{\smallspace}){9-10} 		\cmidrule(l{\tinyspace}r{\smallspace}){12-13}
		&& {E2VID} & {~Ours} && {E2VID} & {~Ours} && {E2VID} & {~Ours} && {EVFlow} & {~Ours} \\
		\midrule
		\midrule
		&\multicolumn{12}{c}{\textbf{HQF}} \\
bike\_bay\_hdr && 0.1580 & \bfseries 0.0299 && 0.4141 & \bfseries 0.5202 && 0.5100 & \bfseries 0.3038			&& 1.2194 & \bfseries 1.2302 \\ 
boxes && 0.1067 & \bfseries 0.0345 && 0.4974 & \bfseries 0.5923 && 0.3755 & \bfseries 0.2575					&& 1.7485 & \bfseries 1.8020 \\ 
desk\_6k && 0.1515 & \bfseries 0.0300 && 0.5117 & \bfseries 0.5966 && 0.3886 & \bfseries 0.2213					&& 1.2250 & \bfseries 1.3515 \\ 
desk\_fast && 0.1151 & \bfseries 0.0354 && 0.5399 & \bfseries 0.6062 && 0.3957 & \bfseries 0.2504				&& 1.4323 & \bfseries 1.4956 \\ 
desk\_hand\_only && 0.1242 & \bfseries 0.0480 && 0.5304 & \bfseries 0.5709 && 0.6319 & \bfseries 0.3864			&& \bfseries 0.9469 & 0.8472 \\ 
desk\_slow && 0.1569 & \bfseries 0.0410 && 0.5306 & \bfseries 0.6218 && 0.4724 & \bfseries 0.2480				&& 1.0085 & \bfseries 1.0756 \\ 
engineering\_posters && 0.1306 & \bfseries 0.0300 && 0.4203 & \bfseries 0.5710 && 0.4680 & \bfseries 0.2560		&& 1.4958 & \bfseries 1.6479 \\ 
high\_texture\_plants && 0.1566 & \bfseries 0.0314 && 0.3686 & \bfseries 0.6476 && 0.3780 & \bfseries 0.1392	&& 0.1283 & \bfseries 1.6809 \\ 
poster\_pillar\_1 && 0.1393 & \bfseries 0.0318 && 0.3803 & \bfseries 0.4990 && 0.5402 & \bfseries 0.2672		&& 1.2026 & \bfseries 1.2413 \\ 
poster\_pillar\_2 && 0.1521 & \bfseries 0.0350 && 0.4009 & \bfseries 0.4745 && 0.5647 & \bfseries 0.2625		&& \bfseries 1.1621 & 0.9649 \\ 
reflective\_materials && 0.1251 & \bfseries 0.0334 && 0.4414 & \bfseries 0.5544 && 0.4436 & \bfseries 0.2802	&& 1.4543 & \bfseries 1.5748 \\ 
slow\_and\_fast\_desk && 0.1620 & \bfseries 0.0286 && 0.4773 & \bfseries 0.6237 && 0.4451 & \bfseries 0.2475	&& 0.9308 & \bfseries 0.9893 \\ 
slow\_hand && 0.1789 & \bfseries 0.0375 && 0.4123 & \bfseries 0.5652 && 0.5701 & \bfseries 0.3023				&& \bfseries 1.6353 & 1.5618 \\ 
still\_life && 0.0940 & \bfseries 0.0261 && 0.5091 & \bfseries 0.6263 && 0.3520 & \bfseries 0.2243				&& 1.9300 & \bfseries 1.9815 \\ 
\midrule
Mean && 0.1448 & \bfseries 0.0326 && 0.4584 & \bfseries 0.5791 && 0.4609 & \bfseries 0.2562 && 1.2045 & \bfseries 1.3540 \\
		\midrule
		\midrule
		&\multicolumn{12}{c}{\textbf{IJRR}} \\
boxes\_6dof\_cut && \bfseries 0.0406 & 0.0419 && 0.6304 & \bfseries 0.6392 && 0.2898 & \bfseries 0.2479		&& 1.4171 & \bfseries 1.4571 \\ 
calibration\_cut && 0.0719 & \bfseries 0.0315 && 0.6119 & \bfseries 0.6245 && 0.2164 & \bfseries 0.1805		&& 1.2040 & \bfseries 1.3057 \\ 
dynamic\_6dof\_cut && 0.1679 & \bfseries 0.0525 && 0.4479 & \bfseries 0.5275 && 0.3750 & \bfseries 0.2673	&& 1.3693 & \bfseries 1.3922 \\ 
office\_zigzag\_cut && 0.0689 & \bfseries 0.0369 && 0.4872 & \bfseries 0.5082 && 0.3086 & \bfseries 0.2599	&& \bfseries 1.1302 & 1.1127 \\ 
poster\_6dof\_cut && 0.0719 & \bfseries 0.0307 && 0.5964 & \bfseries 0.6567 && 0.2559 & \bfseries 0.1947	&& 1.5029 & \bfseries 1.5551 \\ 
shapes\_6dof\_cut && 0.0277 & \bfseries 0.0168 && \bfseries 0.7982 & 0.7652 && 0.2632 & \bfseries 0.2234	&& 1.1485 & \bfseries 1.5699 \\ 
slider\_depth\_cut && 0.0771 & \bfseries 0.0308 && 0.5413 & \bfseries 0.6240 && 0.3459 & \bfseries 0.2434	&& 1.7322 & \bfseries 2.1723 \\ 
\midrule
		Mean && 0.0748 & \bfseries 0.0344 && 0.6102 & \bfseries 0.6364 && 0.2813 & \bfseries 0.2240 && 1.3240 & \bfseries 1.4545 \\
		\midrule
		\midrule
		&\multicolumn{12}{c}{\textbf{MVSEC}} \\
indoor\_flying1\_data\_cut && 0.2467 & \bfseries 0.0840 && 0.1865 & \bfseries 0.3635 && 0.7245 & \bfseries 0.4534 	&& 1.0183 & \bfseries 1.1380\\
indoor\_flying2\_data\_cut && 0.2306 & \bfseries 0.0931 && 0.1821 & \bfseries 0.3581 && 0.7092 & \bfseries 0.4530 	&& 1.1293 & \bfseries 1.3592\\
indoor\_flying3\_data\_cut && 0.2489 & \bfseries 0.0896 && 0.1798 & \bfseries 0.3650 && 0.7343 & \bfseries 0.4440 	&& 1.0569 & \bfseries 1.2291\\
indoor\_flying4\_data\_cut && 0.2107 & \bfseries 0.0784 && 0.2274 & \bfseries 0.3622 && 0.7165 & \bfseries 0.4475 	&& 1.2377 & \bfseries 1.4999\\
outdoor\_day1\_data\_cut && 0.3222 & \bfseries 0.1274 && 0.3096 & \bfseries 0.3395 && 0.6559 & \bfseries 0.5179 	&& 1.1506 & \bfseries 1.2719\\
outdoor\_day2\_data\_cut$^*$ && 0.2968 & \bfseries 0.0963 && 0.2936 & \bfseries 0.3394 && 0.5703 & \bfseries 0.4285 && \bfseries 1.2149 & 1.1958\\
\midrule
		Mean && 0.2895 & \bfseries 0.1052 && 0.2675 & \bfseries 0.3461 && 0.6457 & \bfseries 0.4670 && 1.11856 & \bfseries 1.29962 \\
		\bottomrule
	\end{tabular}
	\begin{tablenotes}
      \item \footnotesize{*Removed from mean tally for EV-FlowNet, as this sequence is part of the training set.}
    \end{tablenotes}
\end{threeparttable}
}
	\label{tab:eval:all_in_one}
\end{table}

\subsubsection{Image}
As in \cite{Rebecq20pami} we compared our reconstructed images to ground truth (DAVIS frames) on three metrics; mean squared error (MSE), structural similarity \cite{Wang04tip} (SSIM) and perceptual loss \cite{Zhang18cvpr} (LPIPS) that uses distance in the latent space of a pretrained deep network to quantify image similarity.

Since many of these datasets show scenes that are challenging for conventional cameras, we carefully selected sections of those sequences where frames appeared to be of higher quality (less blurred, better exposure etc.).
The exact cut times of the IJRR and MVSEC sequences can be found in the supplementary materials.
However, we were also ultimately motivated to record our own dataset of high quality frames (HQF, \seclab{\ref{sec:hqf}}) of which we evaluated the entire sequence.

\subsubsection{Flow}

\label{ssec:flow}
A warping loss (similar to \cite{Gallego17ral}) was used as a proxy measure of accuracy as it doesn't require ground truth flow.
Events \mbox{$E = (x_i,y_i,t_i,s_i)_{i=1,...,N}$} are warped by per-pixel optical flow \mbox{$\phi = (u(x, y), v(x, y))^T$} to a reference time $t^\prime$ via
\begin{equation}
I(E, \phi) = 
\begin{pmatrix}
x_i^\prime\\
y_i^\prime
\end{pmatrix}=
\begin{pmatrix}
x_i\\
y_i
\end{pmatrix}+
(t^\prime-t_i)
\begin{pmatrix}
u(x_i,y_i)\\
v(x_i,y_i)
\end{pmatrix}.
\end{equation}
The resulting image $I$ becomes sharper if the flow is correct, as events are motion compensated.
Sharpness can be evaluated using the variance of the image $\sigma^2(I)$ \cite{Gallego19cvpr, Stoffregen19cvpr}, where a higher value indicates a better flow estimate.
Since image variance $\sigma^2(I)$ depends on scene structure and camera parameters, we normalize by the variance of the unwarped event image $I(E, 0)$ to obtain the \textit{Flow Warp Loss ($\fwl$)}:
\begin{equation}
\fwl := \frac{\sigma^2(I(E, \phi))}{\sigma^2(I(E, 0))}.
\label{eq:fwl}
\end{equation}
$\fwl < 1$ implies the flow is worse than a baseline of zero flow.
$\fwl$ enables evaluation on datasets without ground truth optic flow.
While we used ground truth from the simulator during training, we evaluated on real data using $\fwl$ (\tablab{\ref{tab:eval:all_in_one}}).
We believe training on ground truth (L1 loss) rather than $\fwl$ encourages dense flow predictions.

\tablab{\ref{tab:mvsec_aee}} shows average endpoint error (AEE) of optic flow on MVSEC \cite{Zhu18ral}. 
MVSEC provides optic flow estimates computed from lidar depth and ego motion sensors as `ground truth', allowing us to evaluate average endpoint error (AEE) using code provided in \cite{Zhu18rss}.
However, lidar + ego motion derived ground truth is subject to sensor noise, thus, AEE may be an unreliable metric on MVSEC.
For example, predicting zero flow achieves near state-of-the-art in some cases on MVSEC using AEE, though not with our proposed metric $\fwl$ (by construction, predicting zero flow yields $\fwl=1.0$).

\begin{table}[t]
		\caption{
		Comparison of various methods to optic flow estimated from Lidar depth and ego-motion sensors \cite{Zhu18ral}.
		The average-endpoint-error to the Lidar estimate (AEE) and the percentage of pixels with AEE above 3 and greater than \SI{5}{\percent} of the magnitude of the flow vector (\%$_{\text{Outlier}}$) are presented for each method (lower is better, best in bold).
Zeros shows the baseline error of zero flow.
Additional works are compared in \tablab 9 which can be found in the supplementary materials.
	}
\resizebox{\textwidth}{!}{%
\begin{threeparttable}
	\newcolumntype{Z}{S[table-format=2.2,table-auto-round]}
	\centering
	\setlength{\tabcolsep}{0.6mm}
	\renewcommand*{\arraystretch}{1.1}
	\newcommand{\smallspace}{2mm}
	\small

	\begin{tabular}{@{}lcZZcZZcZZcZZcZZcr@{}}
		\toprule
		\multirow{2}[3]{*}{Dataset} && \multicolumn{2}{c}{outdoor\_day1} && \multicolumn{2}{c}{outdoor\_day2} && \multicolumn{2}{c}{indoor\_flying1}  && \multicolumn{2}{c}{indoor\_flying2}  && \multicolumn{2}{c}{indoor\_flying3} \\
		\cmidrule(l{\smallspace}r{\smallspace}){3-4} \cmidrule(l{\smallspace}r{\smallspace}){6-7} \cmidrule(l{\smallspace}r{\smallspace}){9-10} \cmidrule(l{\smallspace}r{\smallspace}){12-13} \cmidrule(l{\smallspace}r{\smallspace}){15-16}
		&& {AEE} & {\%$_{\text{Outlier}}$} && {AEE} & {\%$_{\text{Outlier}}$} && {AEE} & {\%$_{\text{Outlier}}$} && {AEE} & {\%$_{\text{Outlier}}$} && {AEE} & {\%$_{\text{Outlier}}$} \\
		\midrule
Zeros && 4.31 & 0.39 && 1.07 & \bfseries 0.91 && 1.10 & \bfseries 1.00 && 1.74 & \bfseries 0.89 && 1.50 & \bfseries  0.94 \\
EVFlow \cite{Zhu18rss} && \bfseries 0.49 & \bfseries 0.2 && $\text{-}$ & $\text{-}$ && 1.03 &  2.20 && 1.72 & 15.10 && 1.53 & 11.90 \\
Ours && 0.68 & 0.99 && \bfseries 0.82 & 0.96 && \bfseries 0.56 & \bfseries 1.00 && \bfseries 0.66 & 1.00 && \bfseries 0.59 & 1.00 \\
		\bottomrule
	\end{tabular}
\end{threeparttable}
}
	\label{tab:mvsec_aee}
\end{table}

\subsection{Contrast Thresholds}
\label{sec:ct_experiments}
We investigated the impact of simulator contrast threshold ($\cth$, see \seclab{\ref{sec:choice_of_ct}}) by retraining several networks on simulated datasets with $\cth$s ranging from 0.2 to 1.5.
Each dataset contained the same sequences, differing only in $\cth$.
\tablab{\ref{tab:ct_experiment}} shows that for reconstruction (evaluated on LPIPS), IJRR is best on a lower $\cth\approx0.2$, while MVSEC is best on high $\cth\approx1.0$.
Best or runner up performance was achieved when a wide range of $\cth$s was used, indicating that exposing a network to additional event statistics outside the inference domain is not harmful, and may be beneficial.
We believe training with low $\cth$s (thus higher $\eppps$) reduces dynamic range in the output images (\tablab{\ref{tab:e2vid_hdr_comparison}}), perhaps because the network becomes accustomed to a high density of events during training but is presented with lower $\eppps$ data at inference.
When retraining the original E2VID network, dynamic range increases with $\cth$s (\tablab{\ref{tab:e2vid_hdr_comparison}}).

\begin{table}[t]
	\caption{
		Evaluation of image reconstruction and optic flow networks trained on simulated datasets with a variety of contrast thresholds ($\cth$s) from 0.2 to 1.5.
		`All' is a dataset containing the full range of $\cth$s from 0.2 to 1.5.
		All networks are trained for 200 epochs and evaluated on datasets HQF (excluding \dsfont{desk\_hand\_only} on $\fwl$), IJRR \cite{Mueggler17ijrr}, MVSEC \cite{Zhu18ral}.
		We report mean squared error (MSE), structural similarity (SSIM) \cite{Wang04tip} and perceptual loss (LPIPS) \cite{Zhang18cvprLPIPS} for reconstruction and $\fwl$ for optic flow. 
		Key: \textbf{best} $|$ \textit{second best}.
	}
	\newcolumntype{Z}{S[table-format=2.2,table-auto-round]}
	\centering
	\setlength{\tabcolsep}{0.8mm}
	\renewcommand*{\arraystretch}{1.1}
	\newcommand{\smallspace}{2mm}
	\sisetup{detect-all = true}
	\small
	\resizebox{\textwidth}{!}{%
	\begin{tabular}{@{}c c ZZZcZ c ZZZcZ c ZZZcZ cr@{}}
		\toprule
		\multirow{2}[3]{8ex}{Contrast\\threshold} && \multicolumn{5}{c}{HQF} && \multicolumn{5}{c}{IJRR} && \multicolumn{5}{c}{MVSEC}
		\tabularnewline
		\cmidrule(l{\smallspace}r{\smallspace}){3-7} \cmidrule(l{\smallspace}r{\smallspace}){9-13} \cmidrule(l{\smallspace}r{\smallspace}){15-19}
		&& {MSE} & {SSIM} & {LPIPS} && {\textit{FWL}} && {MSE} & {SSIM} & {LPIPS} && {\textit{FWL}} && {MSE} & {SSIM} & {LPIPS} && {FWL}
		\tabularnewline
		\midrule
0.20
&& 0.05 & 0.50 & 0.38 && 1.9258
&& \bfseries 0.04 & \bfseries 0.60 & \bfseries 0.25 && \itshape 1.453
&& 0.10 & \bfseries 0.35 & 0.55 && 1.1525
\tabularnewline
0.50
&& \bfseries 0.04 & \itshape 0.51 & 0.36 && 1.8996
&& 0.04 & 0.57 & 0.27 && 1.424
&& \itshape 0.10 & 0.31 & 0.52 && 1.1850
\tabularnewline
0.75
&& 0.05 & \bfseries 0.51 & \bfseries 0.36 && 1.9035 
&& 0.05 & 0.56 & 0.28 && 1.438
&& 0.11 & 0.29 & 0.53 && \itshape 1.2238
\tabularnewline
1.00
&& 0.05 & 0.48 & 0.36 && 1.9093
&& 0.05 & 0.53 & 0.29 && 1.419
&& 0.12 & 0.27 & \itshape 0.51 && 1.1831
\tabularnewline
1.50
&& 0.05 & 0.47 & 0.38 && \itshape 1.9272
&& 0.06 & 0.52 & 0.30 && 1.436
&& 0.09 & 0.30 & 0.52 && 1.1431
\tabularnewline
All
&& \itshape 0.05 & 0.50 & \bfseries 0.36 && \bfseries 1.9617
&& \bfseries 0.04 & \itshape 0.59 & \itshape 0.27 && \bfseries 1.459
&& \bfseries 0.08 & \itshape 0.34 & \bfseries 0.51 && \bfseries 1.2434
\tabularnewline
\bottomrule
 \end{tabular}
}
\label{tab:ct_experiment}
\end{table}

\begin{table}[t]
	\caption{Dynamic range of reconstructed images from IJRR \cite{Mueggler17ijrr}: original E2VID \cite{Rebecq20pami} versus E2VID retrained on simulated datasets covering a range of contrast thresholds $\cth$s.
	We report the mean dynamic range of the 10th-90th percentile of pixel values.
	}
	\label{tab:e2vid_hdr_comparison}
	\centering
	\newcommand{\smallspace}{3mm}
	\renewcommand*{\arraystretch}{1.1}
	\begin{tabular}{lSSSSSSS}
		\toprule
		& {Original \cite{Rebecq20pami}} & \multicolumn{6}{c}{Retrained}
		\tabularnewline
		\cmidrule(l{\smallspace}r{\smallspace}){2-2} 		\cmidrule(l{\smallspace}r{\smallspace}){3-8} 
		{Contrast threshold} & \sim 0.18 & 0.2 & 0.5 & 0.75 & 1.0 & 1.5 &
		All
		\tabularnewline
		{Dynamic range} & 77.3 & 89.2 & 103.7 & 105.9 & 104.8 & 100.0 & 103.3
		\tabularnewline
		\bottomrule
	\end{tabular}
\end{table}

\subsection{Training Noise and Sequence Length}

To determine the impact of sequence length and noise augmentation during training, we retrained E2VID architecture using sequence length 40 (L40) and 120 (L120), with and without noise augmentation (N) (see \tablab{\ref{tab:ablation_all}}).
Increasing sequence length from 40 to 120 didn't impact results significantly.
Noise augmentation during training improved performance of L40 models by $\sim$ \SI{5}{}-\SI{10}{\percent}, while giving mixed results on different datasets for L120 models.
Qualitatively, adding more noise encourages networks to smooth outputs, while less noise may encourage the network to `reconstruct' noise events, resulting in artifacts (\figlab{\ref{fig:front_page}}) observed in E2VID \cite{Rebecq20pami} (trained without noise).

\begin{table}
		\caption{
		Mean LPIPS \cite{Zhang18cvprLPIPS} on our HQF dataset, IJRR \cite{Mueggler17ijrr} and MVSEC \cite{Zhu18ral}, for various training hyperparameter configurations. E2VID architecture retrained from scratch in all experiments. Key: L40/L120=sequence length 40/120, N=noise augmentation during training.}
	\newcolumntype{Z}{S[table-format=2.3,table-auto-round]}
	\centering
	\setlength{\tabcolsep}{0.8mm}
	\renewcommand*{\arraystretch}{1.1}
	\newcommand{\smallspace}{3mm}
	\small
	\resizebox{\textwidth}{!}{%
	\begin{tabular}{@{}lcZZZcZZZcZZZcr@{}}
		\toprule
		\multirow{2}[3]{*}{Model} && \multicolumn{3}{c}{HQF} && \multicolumn{3}{c}{IJRR} && \multicolumn{3}{c}{MVSEC} \\
		\cmidrule(l{\smallspace}r{\smallspace}){3-5} \cmidrule(l{\smallspace}r{\smallspace}){7-9} \cmidrule(l{\smallspace}r{\smallspace}){11-13}
		&& {MSE} & {SSIM} & {LPIPS} && {MSE} & {SSIM} & {LPIPS} && {MSE} & {SSIM} & {LPIPS} \\
		\midrule
		L40
		&& 0.0441 & \bfseries 0.5826 & 0.2964
		&& 0.0417 & \bfseries 0.6497 & 0.2289
		&& 0.1514 & 0.3302 & 0.5255 \\
		L40N
		&& \bfseries 0.0326 & 0.5791 & \bfseries 0.2562
		&& \bfseries 0.0344 & 0.6364 & \bfseries 0.2240
		&& 0.1052 & \bfseries 0.3461 & \bfseries 0.4670 \\
		L120
		&& 0.0399 & 0.5444 & 0.2791
		&& 0.0383 & 0.6187 & 0.2368
		&& 0.1321 & 0.3112 & 0.4775 \\
		L120N
		&& 0.0358 & 0.5468 & 0.2904
		&& 0.0396 & 0.6079 & 0.2410
		&& \bfseries 0.0990 & 0.3439 & 0.4983 \\
		\bottomrule
	\end{tabular}
}
	\label{tab:ablation_all}
\end{table}

\section{Discussion}
\label{sec:discussion}
The significant improvements gained by training models on our synthetic dataset exemplify the importance of reducing the sim-to-real gap for event cameras in both the event rate induced by varying the contrast thresholds and the dynamics of the simulation scenes.
Our results are quite clear on this, with consistent improvements across tasks (reconstruction and optic flow) and architectures (recurrent networks like E2VID, and U-Net based flow estimators) of up to \SI{40}{\percent}.

We believe this highlights the importance for researchers to pay attention to the properties of the events they are training on; are the settings of the camera or simulator such that they are generating more or less events?
Are the scenes they are recording representative of the wide range of scenes that are likely to be encountered during inference?

In particular, it seems that previous works have inadvertently overfit their models to the events found in the chosen target dataset.
EV-FlowNet performs better on sequences whose dynamics are similar to the slow, steady scenes in MVSEC used for training, examples being \dsfont{poster\_pillar\_2} or \dsfont{desk\_slow} from HQF that feature long pauses and slow motions, where EV-FlowNet is on par or better than ours.
For researchers looking to use an off-the-shelf pretrained network, our model may be a better fit, since it targets a greater variety of sensors and scenes.
A further advantage of our model that is not reflected in the $\fwl$ metric, is that training in simulation allows our model to predict \textit{dense} flow (see supp. material), a challenge for prior self-supervised methods.

Similarly, our results speak for themselves on image reconstruction.
While we outperform E2VID \cite{Rebecq20pami} on all datasets, the smallest gap is on IJRR, the dataset we found to have lower $\cth$s.
E2VID performs worst on MVSEC that contains higher $\cth$s, consistent with our finding that performance is driven by similarity between training and evaluation event data. 

In conclusion, future networks trained with synthetic data from ESIM or other simulators should take care to ensure the statistics of their synthetic data match the final use-case, using large ranges of $\cth$ values and appropriate noise and pause augmentation in order to ensure generalized models.

\section*{Acknowledgments}
This work was supported by the Australian Government Research Training Program Scholarship and the Australian Research Council through the ``Australian Centre of Excellence for Robotic Vision'' under Grant CE140100016.

\FloatBarrier
\clearpage
{\small
\bibliographystyle{ieee_fullname}
\bibliography{ced_bib/all.bib}
}
\clearpage
\section{Supplementary Materials}
\subsection{HQF Dataset Details}
The details of the High Quality Frames dataset are shown in \tablab \ref{tab:HQFD_description}.
\begin{table}
	\centering
\caption{Breakdown of the sequences included in HQF.
To provide some inter-device variability, the dataset is taken with two separate DAVIS 240C cameras, 1 and 2.}
\label{tab:HQFD_description}
\newcolumntype{Z}{S[table-format=2.1,table-auto-round]}
\renewcommand\multirowsetup{\centering}
\renewcommand*{\arraystretch}{1.1}
\small
\resizebox{\textwidth}{!}{
\begin{tabular}{lZZZZl}
\toprule
{\multirow{2}{*}{Sequence}} &
{\multirow{2}{8ex}{Length\\{[s]}}} &
{\multirow{2}{*}{Cam.}} & 
{\multirow{2}{8ex}{Frames\\{[k]}}} &
{\multirow{2}{8ex}{Events\\{[M]}}} &
{\multirow{2}{*}{Description}}
\tabularnewline
\tabularnewline 
\midrule
bike\_bay\_hdr & 99.0 & 1 & 2.431 & 19.763043 & 		Camera moves from dim to bright\tabularnewline
boxes & 24.2 & 1 & 0.540 & 10.050781 & 				Indoor light, translations\tabularnewline
desk & 65.8 & 2 & 1.491 & 13.516351 & 				Natural light, various motions\tabularnewline
desk\_fast & 32.0 & 2 & 0.724 & 12.622934 & 			Natural light, fast motions\tabularnewline
desk\_hand\_only & 20.6 & 2 & 0.467 & 0.777177 & 		Indoor light, static camera\tabularnewline
desk\_slow & 63.3 & 2 & 1.434 & 1.926605 & 			Natural light, slow motions\tabularnewline
engineering\_posters & 60.7 & 1 & 1.267 & 15.376272 & Indoor light, text and images\tabularnewline
high\_texture\_plants & 43.2 & 1 & 1.090 & 14.602389 &Outdoors, high textures\tabularnewline
poster\_pillar\_1 & 41.8 & 1 & 0.998 & 7.056573 & 		Outdoors, text and images\tabularnewline
poster\_pillar\_2 & 25.4 & 1 & 0.613 & 2.540922 & 		Outdoors, text and images, long pause\tabularnewline
reflective\_materials & 28.9 & 1 & 0.619 & 7.753856 & 	Natural light, reflective objects\tabularnewline
slow\_and\_fast\_desk & 75.6 & 1 & 1.744 & 15.030414 &Natural light, diverse motion\tabularnewline
slow\_hand & 38.9 & 1 & 0.901 & 7.635825 & 			Indoor, slow motion, static camera\tabularnewline
still\_life & 68.1 & 1 & 1.194 & 42.740481 & Indoors, Indoor light, 6DOF motions\tabularnewline
\bottomrule
\end{tabular}
}
\end{table}

\subsection{Voxel Generation}
\begin{figure}
\global\long\def\voxel_generation_width{0.9\linewidth}
\centering
\includegraphics[width=\voxel_generation_width]{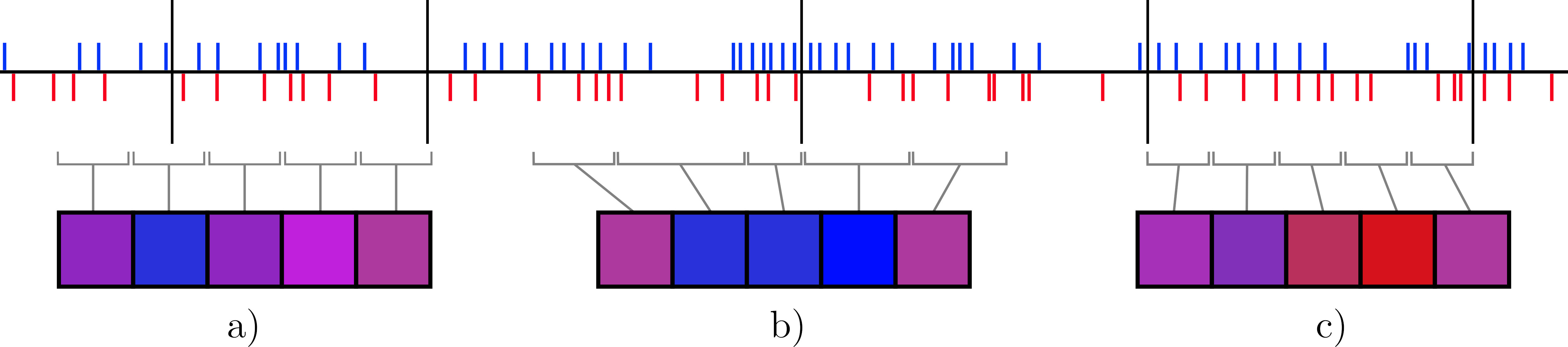}
    \caption{\label{fig:voxel_generation} 
    Events (blue and red lines) on a timeline are discretized into voxels below (squares) according to: a) \textit{fixed rate}, b) \textit{fixed events}, c) \textit{between frames} (frames denoted by black lines).
}
\end{figure}

Two natural choices for generating voxel grids from the event stream are \textit{fixed rate} and \textit{fixed events} (\figlab{\ref{fig:voxel_generation}}). 
In fixed rate, voxels are formed from $t$ second wide slices of the event stream (variable event count), endowing the resulting inference with a fixed frame rate. 
This has the downside that inference cannot adapt to changing scene dynamics, a disadvantage shared by conventional cameras.
A special case of fixed rate is \textit{between frames} where all events between two image frames are used to form a voxel grid.

In fixed events, one waits for $N$ events before making a voxel grid no matter how long it takes (variable duration).
Fixed events has the downside that if the camera receives few events, either because the scene has little texture or the motion is slow, the inference rate can slow to a crawl.
This method allows matching the value $N$ to the average $N$ of the training set during inference, potentially benefiting the network.
The average events per voxel in our training set is $0.0564$.

\subsection{CTs extended}
As outlined in \seclab \ref{sec:method}, we propose several methods for estimating the $\cth$s for a given event-based dataset.
In total, we tried three different methods:
\begin{itemize}
\item Creating a simulator scene with similar texture and range of motions as the real sequence and adjusting the $\cth$s until the $\eppps$ match.
\item Simulating events from the APS frames (if they are available) and adjusting the $\cth$s until the $\eppps$ match the real sequence.
\item Creating a calibration scene in the simulator, recording this scene on-screen and adjusting the $\cth$ of the event camera to match the $\eppps$ of the simulation.
\end{itemize}
We describe the first two approaches in the main paper.
These approaches indicate $\cth$s of approximately $0.3$ and $0.75$ for IJRR and MVSEC respectively.

We also produced a calibration sequence in an attempt to match the simulator to our particular DAVIS 240C at default settings.
For this, we moved a checkerboard across the image plane in ESIM, using various $\cth$s.
The scene was played on a high-refresh screen and recorded by our DAVIS.
The resulting event-rate for each sequence, shown in \tablab \ref{tab:checkerboard_calibration}, suggest a $\cth\approx0.5$ to match the camera to the simulator.
This is however in conflict with the $\eppps$ ($8.2$) of the real sequences compared to the $\eppps$ of the best training data $\cth$ ($19.6$).
In other words, there seems to be a mismatch for this method of calibration, perhaps stemming from a difference in recording events from real scenes to recording scenes from a screen as was done for the checkerboard calibration sequence.

\begin{table}
	\caption{Comparison of the $\eppps$ for simulated sequences at various $\cth$ settings with the $\eppps$ of a real calibration sequence.
	The sequence consists of a checkerboard in motion.
	The same sequence is also recorded by a real event camera (DAVIS 240C) using default bias settings.
	The result suggests that a $\cth$ value around 0.5 would be appropriate to match the simulator to the real camera.}
	\centering
	\setlength{\tabcolsep}{1.4mm}
	\renewcommand*{\arraystretch}{1.1}
	\begin{tabular}{l SSSSSS}
		\toprule
		\makecell[l]{Contrast\\threshold} & {0.2} & {0.5} & {0.75} & {1.0} & {1.5} & {Real}
		\tabularnewline
		\midrule
		$\eppps$ \smallskip & 26.5 & 19.6 & 16.2 & 10.8 & 7.2 & 8.2
		\tabularnewline
		LPIPS & 0.289 & 0.285 & 0.289 & 0.311 & 0.316 & $\text{-}$
		\tabularnewline
		\bottomrule
	\end{tabular}
	\label{tab:checkerboard_calibration}
\end{table}

\newpage

\subsection{Sequence Cuts}
Since the frames accompanying the events in the commonly used MVSEC and IJRR datasets are of low quality, we only evaluate on select sequences and for select cuts of those sequences.
These cuts are enumerated in \tablab \ref{tab:dataset_cuts}.

\begin{table}
\caption{Start and end times for sequences in IJRR and MVSEC that we present validation statistics on.
While both IJRR and MVSEC contain more sequences than the ones listed, those not included had very low quality accompanying frames (see \figlab \ref{fig:dataset_quality_comparison}).}
\centering
\newcommand{\smallspace}{2mm}
\setlength{\tabcolsep}{1.4mm}
\renewcommand*{\arraystretch}{1.1}
\begin{tabular}{lSSclSS}
\toprule
\multicolumn{3}{c}{{IJRR}} && \multicolumn{3}{c}{{MVSEC}}
\tabularnewline
\cmidrule(l{\smallspace}r{\smallspace}){1-3} \cmidrule(l{\smallspace}r{\smallspace}){5-7}
{Sequence} & {Start [s]} & {End [s]} && {Sequence} & {Start [s]} & {End [s]}
\tabularnewline
\midrule
boxes\_6dof & 5.0 & 20.0 && indoor\_flying1 & 10.0 & 70.0
\tabularnewline
calibration & 5.0 & 20.0 && indoor\_flying2 & 10.0 & 70.0
\tabularnewline
dynamic\_6dof & 5.0 & 20.0 && indoor\_flying3 & 10.0 & 70.0
\tabularnewline
office\_zigzag & 5.0 & 12.0 && indoor\_flying4 & 10.0 & 19.8
\tabularnewline
poster\_6dof & 5.0 & 20.0 && outdoor\_day1 & 0.0 & 60.0
\tabularnewline
shapes\_6dof & 5.0 & 20.0 && outdoor\_day2 & 100.0 & 160.0
\tabularnewline
slider\_depth & 1.0 & 2.5 &&  &  &
\tabularnewline
\bottomrule
\end{tabular}
\label{tab:dataset_cuts}
\end{table}

\subsection{MVSEC Expanded Results}
For space reasons we did not include results of other works which quote MVSEC.
This is because these works did not release the models, which meant we could not compare on the $\fwl$ metric, so we did not include them in the main paper.
However, for completeness, we show the results of \cite{Ye19arxiv, Zhu19cvpr} and \cite{Gehrig19iccv}.
As can be seen in \tablab \ref{tab:mvsec_aee_all}, our network still compares very favorably.
Interestingly, zero loss (doing nothing) is still the best overall at reducing outliers and is a strong contender for AEE (especially in the flying sequences), showing the importance of reporting the relative improvement as in our $\fwl$.
\begin{table}[t]
		\caption{
		Comparison of various methods to optic flow estimated from Lidar depth and ego-motion sensors \cite{Zhu18ral}.
		The average-endpoint-error to the Lidar estimate (AEE) and the percentage of pixels with AEE above 3 and greater than \SI{5}{\percent} of the magnitude of the flow vector (\%$_{\text{Outlier}}$) are presented for each method (lower is better, best in bold).
		The time between frames is dt=1.
Zeros is the baseline error resulting from always estimating zero flow.
	}
\resizebox{\textwidth}{!}{%
\begin{threeparttable}
	\newcolumntype{Z}{S[table-format=2.2,table-auto-round]}
	\centering
	\setlength{\tabcolsep}{0.6mm}
	\renewcommand*{\arraystretch}{1.1}
	\newcommand{\smallspace}{2mm}
	\small

	\begin{tabular}{@{}lcZZcZZcZZcZZcZZcr@{}}
		\toprule
		\multirow{2}[3]{*}{Dataset} && \multicolumn{2}{c}{outdoor\_day1} && \multicolumn{2}{c}{outdoor\_day2} && \multicolumn{2}{c}{indoor\_flying1}  && \multicolumn{2}{c}{indoor\_flying2}  && \multicolumn{2}{c}{indoor\_flying3} \\
		\cmidrule(l{\smallspace}r{\smallspace}){3-4} \cmidrule(l{\smallspace}r{\smallspace}){6-7} \cmidrule(l{\smallspace}r{\smallspace}){9-10} \cmidrule(l{\smallspace}r{\smallspace}){12-13} \cmidrule(l{\smallspace}r{\smallspace}){15-16}
		&& {AEE} & {\%$_{\text{Outlier}}$} && {AEE} & {\%$_{\text{Outlier}}$} && {AEE} & {\%$_{\text{Outlier}}$} && {AEE} & {\%$_{\text{Outlier}}$} && {AEE} & {\%$_{\text{Outlier}}$} \\
		\midrule
Zeros && 4.31 & 0.39 && 1.07 & \bfseries 0.91 && 1.10 &  1.00 && 1.74 &  \bfseries  0.89 && 1.50 & \bfseries 0.94 \\
EVFlow \cite{Zhu18rss} &&  0.49 &  0.2 && $\text{-}$ & $\text{-}$ && 1.03 &  2.20 && 1.72 & 15.10 && 1.53 & 11.90 \\
EVFlow+ \cite{Zhu19cvpr} && \bfseries 0.32 & \bfseries 0.0 && $\text{-}$ & $\text{-}$ && 0.58 & \bfseries 0.0 && 1.02 & 4.0 && 0.87 & 3.0 \\
Gehrig \cite{Gehrig19iccv} && $\text{-}$ & $\text{-}$ && $\text{-}$ & $\text{-}$ && 0.96 & 0.91 && 1.38 & 8.20 && 1.40 & 6.47 \\
Ours && 0.68 & 0.99 && \bfseries  0.82 & 0.96 &&  \bfseries 0.56 &  1.00 &&  \bfseries 0.66 & 1.00 &&  \bfseries 0.59 & 1.00 \\
\midrule
ECN* \cite{Ye19arxiv} &&  0.35 &  0.04 && $\text{-}$ & $\text{-}$ &&  0.21 &  0.01&& $\text{-}$ & $\text{-}$&& $\text{-}$ & $\text{-}$\tabularnewline
		\bottomrule
	\end{tabular}
	\begin{tablenotes}
      \item \footnotesize{*ECN is trained on \SI{80}{\percent} of the sequence and evaluated on the remaining \SI{20}{\percent}.
      This prevents direct comparison, however we include their result for completeness sake.}
    \end{tablenotes}
\end{threeparttable}
}
	\label{tab:mvsec_aee_all}
\end{table}

\subsection{Additional Qualitative Results}
\def \widthplot {0.25\linewidth}
\begin{longtable}{ccc}
E2VID & Ours & Groundtruth\\
\multicolumn{3}{c}{\dsfont{bike\_bay\_hdr}}\\
\frame{\includegraphics[width=\widthplot]{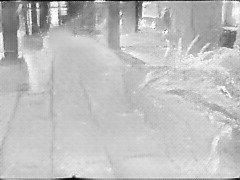}} &
\frame{\includegraphics[width=\widthplot]{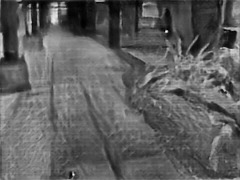}} &
\frame{\includegraphics[width=\widthplot]{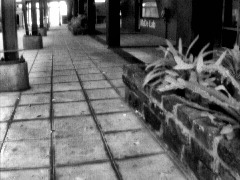}} \\

\multicolumn{3}{c}{\dsfont{boxes}}\\
\frame{\includegraphics[width=\widthplot]{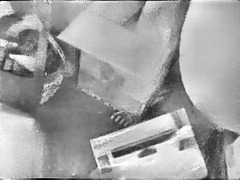}} &
\frame{\includegraphics[width=\widthplot]{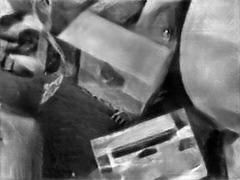}} &
\frame{\includegraphics[width=\widthplot]{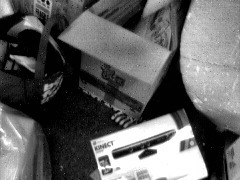}} \\

\multicolumn{3}{c}{\dsfont{desk\_6k}}\\
\frame{\includegraphics[width=\widthplot]{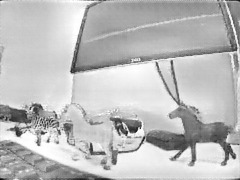}} &
\frame{\includegraphics[width=\widthplot]{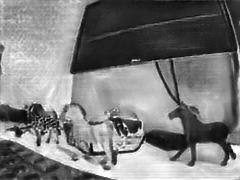}} &
\frame{\includegraphics[width=\widthplot]{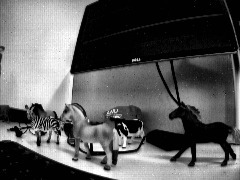}} \\

\multicolumn{3}{c}{\dsfont{desk\_fast}}\\
\frame{\includegraphics[width=\widthplot]{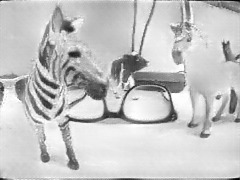}} &
\frame{\includegraphics[width=\widthplot]{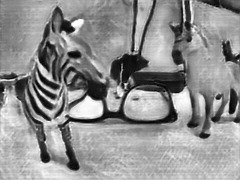}} &
\frame{\includegraphics[width=\widthplot]{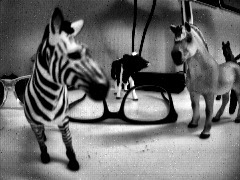}} \\

\multicolumn{3}{c}{\dsfont{desk\_hand\_only}}\\
\frame{\includegraphics[width=\widthplot]{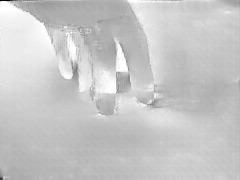}} &
\frame{\includegraphics[width=\widthplot]{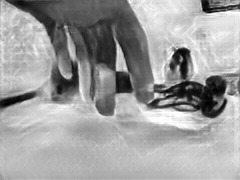}} &
\frame{\includegraphics[width=\widthplot]{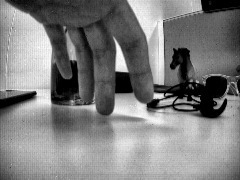}} \\

\multicolumn{3}{c}{\dsfont{desk\_slow}}\\
\frame{\includegraphics[width=\widthplot]{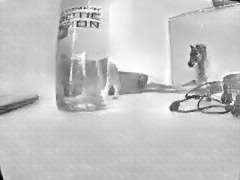}} &
\frame{\includegraphics[width=\widthplot]{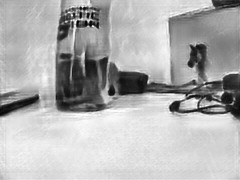}} &
\frame{\includegraphics[width=\widthplot]{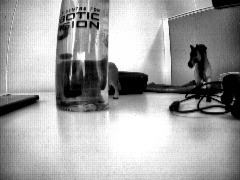}} \\

\multicolumn{3}{c}{\dsfont{engineering\_posters}}\\
\frame{\includegraphics[width=\widthplot]{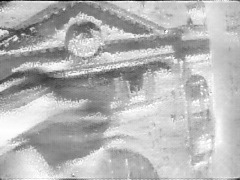}} &
\frame{\includegraphics[width=\widthplot]{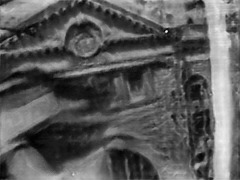}} &
\frame{\includegraphics[width=\widthplot]{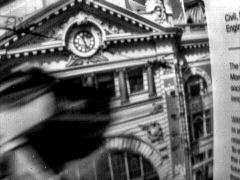}} \\

\multicolumn{3}{c}{\dsfont{high\_texture\_plants}}\\
\frame{\includegraphics[width=\widthplot]{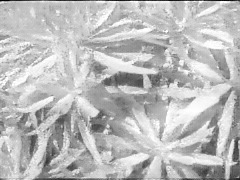}} &
\frame{\includegraphics[width=\widthplot]{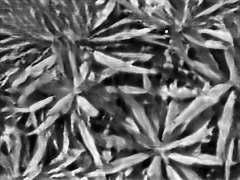}} &
\frame{\includegraphics[width=\widthplot]{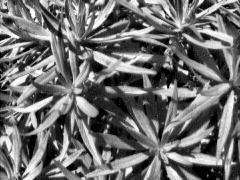}} \\

\multicolumn{3}{c}{\dsfont{poster\_pillar\_1}}\\
\frame{\includegraphics[width=\widthplot]{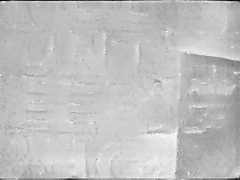}} &
\frame{\includegraphics[width=\widthplot]{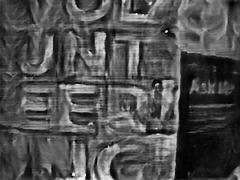}} &
\frame{\includegraphics[width=\widthplot]{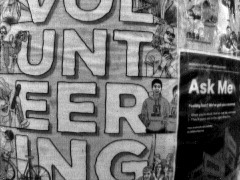}} \\

\multicolumn{3}{c}{\dsfont{poster\_pillar\_2}}\\
\frame{\includegraphics[width=\widthplot]{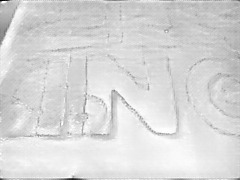}} &
\frame{\includegraphics[width=\widthplot]{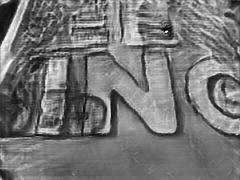}} &
\frame{\includegraphics[width=\widthplot]{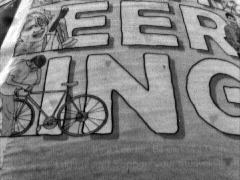}} \\

\multicolumn{3}{c}{\dsfont{reflective\_materials}}\\
\frame{\includegraphics[width=\widthplot]{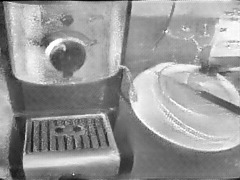}} &
\frame{\includegraphics[width=\widthplot]{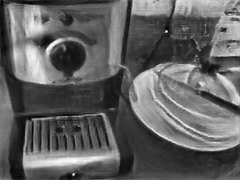}} &
\frame{\includegraphics[width=\widthplot]{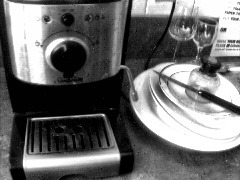}} \\

\multicolumn{3}{c}{\dsfont{slow\_and\_fast\_desk}}\\
\frame{\includegraphics[width=\widthplot]{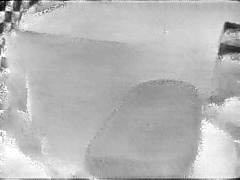}} &
\frame{\includegraphics[width=\widthplot]{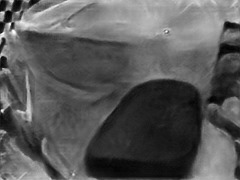}} &
\frame{\includegraphics[width=\widthplot]{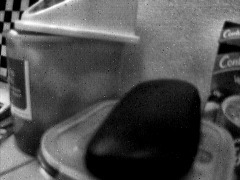}} \\

\multicolumn{3}{c}{\dsfont{slow\_hand}}\\
\frame{\includegraphics[width=\widthplot]{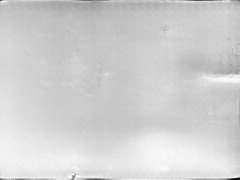}} &
\frame{\includegraphics[width=\widthplot]{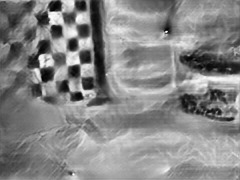}} &
\frame{\includegraphics[width=\widthplot]{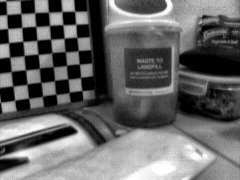}} \\

\multicolumn{3}{c}{\dsfont{still\_life}}\\
\frame{\includegraphics[width=\widthplot]{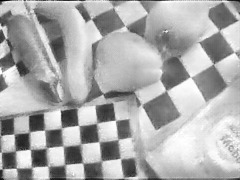}} &
\frame{\includegraphics[width=\widthplot]{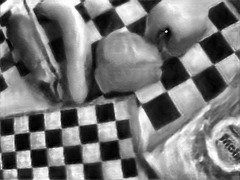}} &
\frame{\includegraphics[width=\widthplot]{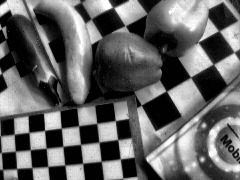}} \\

\caption{Qualitative results for HQFD. Random selection, not cherry picked.}
\end{longtable}

\def \widthplot {0.25\linewidth}
\begin{longtable}{ccc}
E2VID & Ours & Groundtruth\\
\multicolumn{3}{c}{\dsfont{boxes\_6dof\_cut}}\\
\frame{\includegraphics[width=\widthplot]{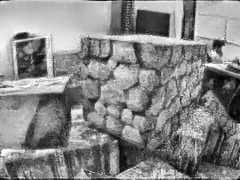}} &
\frame{\includegraphics[width=\widthplot]{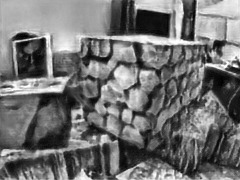}} &
\frame{\includegraphics[width=\widthplot]{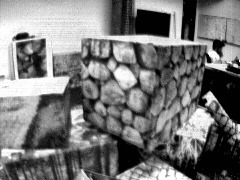}} \\

\multicolumn{3}{c}{\dsfont{calibration\_cut}}\\
\frame{\includegraphics[width=\widthplot]{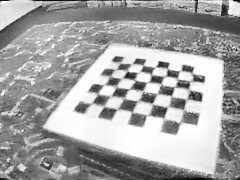}} &
\frame{\includegraphics[width=\widthplot]{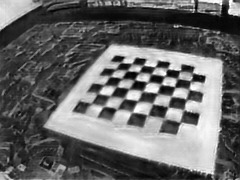}} &
\frame{\includegraphics[width=\widthplot]{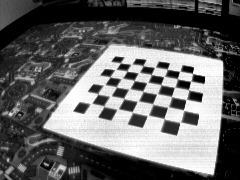}} \\

\multicolumn{3}{c}{\dsfont{dynamic\_6dof\_cut}}\\
\frame{\includegraphics[width=\widthplot]{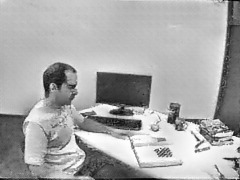}} &
\frame{\includegraphics[width=\widthplot]{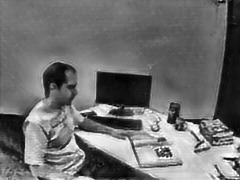}} &
\frame{\includegraphics[width=\widthplot]{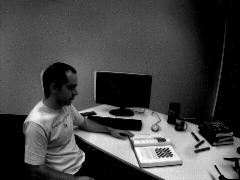}} \\

\multicolumn{3}{c}{\dsfont{office\_zigzag\_cut}}\\
\frame{\includegraphics[width=\widthplot]{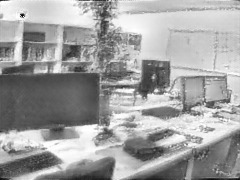}} &
\frame{\includegraphics[width=\widthplot]{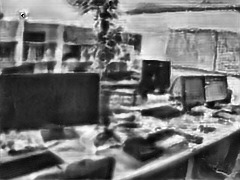}} &
\frame{\includegraphics[width=\widthplot]{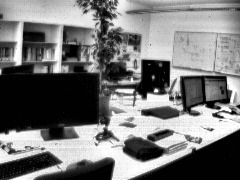}} \\

\multicolumn{3}{c}{\dsfont{poster\_6dof\_cut}}\\
\frame{\includegraphics[width=\widthplot]{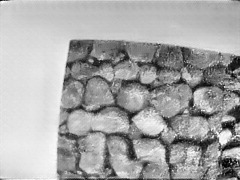}} &
\frame{\includegraphics[width=\widthplot]{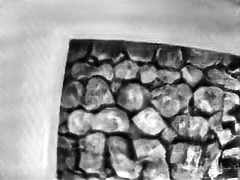}} &
\frame{\includegraphics[width=\widthplot]{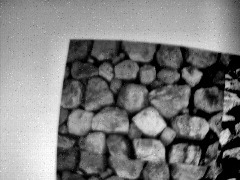}} \\

\multicolumn{3}{c}{\dsfont{shapes\_6dof\_cut}}\\
\frame{\includegraphics[width=\widthplot]{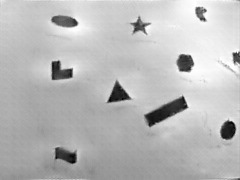}} &
\frame{\includegraphics[width=\widthplot]{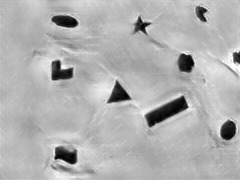}} &
\frame{\includegraphics[width=\widthplot]{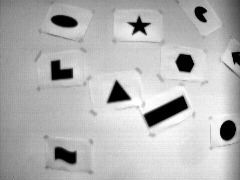}} \\

\multicolumn{3}{c}{\dsfont{slider\_depth\_cut}}\\
\frame{\includegraphics[width=\widthplot]{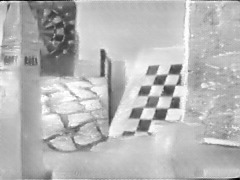}} &
\frame{\includegraphics[width=\widthplot]{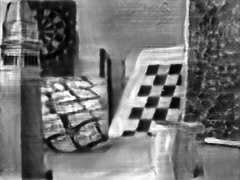}} &
\frame{\includegraphics[width=\widthplot]{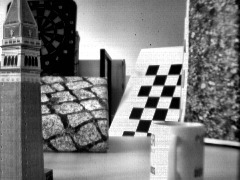}} \\

\caption{Qualitative results for IJRR. Random selection, not cherry picked.}
\end{longtable}

\newpage
\def \widthplot {0.25\linewidth}
\begin{longtable}{ccc}
E2VID & Ours & Groundtruth\\
\multicolumn{3}{c}{\dsfont{indoor\_flying1\_data\_cut}}\\
\frame{\includegraphics[width=\widthplot]{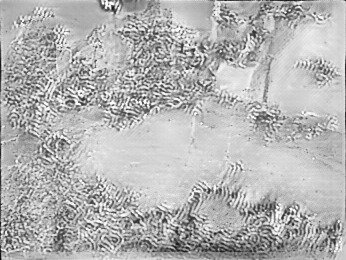}} &
\frame{\includegraphics[width=\widthplot]{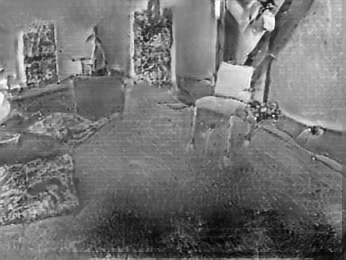}} &
\frame{\includegraphics[width=\widthplot]{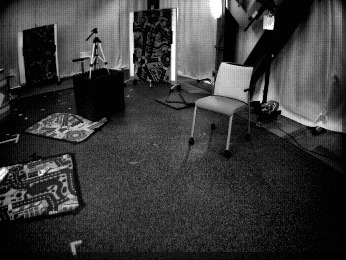}} \\

\multicolumn{3}{c}{\dsfont{indoor\_flying2\_data\_cut}}\\
\frame{\includegraphics[width=\widthplot]{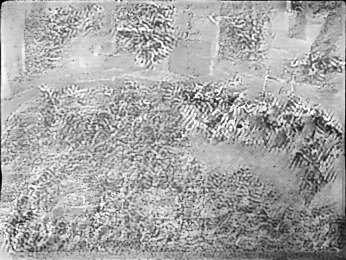}} &
\frame{\includegraphics[width=\widthplot]{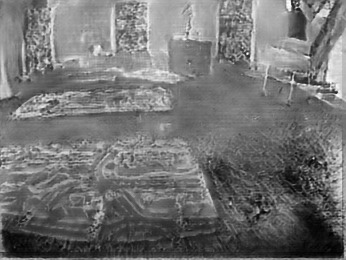}} &
\frame{\includegraphics[width=\widthplot]{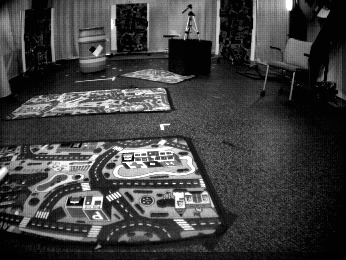}} \\

\multicolumn{3}{c}{\dsfont{indoor\_flying3\_data\_cut}}\\
\frame{\includegraphics[width=\widthplot]{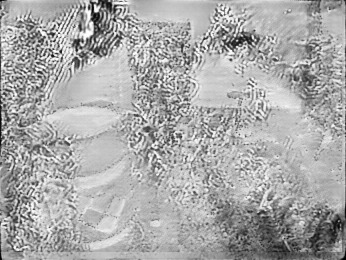}} &
\frame{\includegraphics[width=\widthplot]{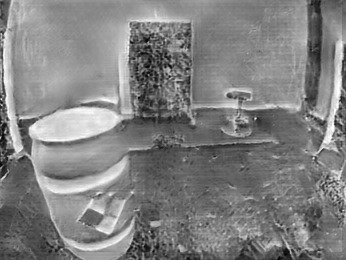}} &
\frame{\includegraphics[width=\widthplot]{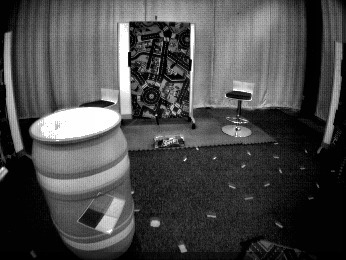}} \\

\multicolumn{3}{c}{\dsfont{indoor\_flying4\_data\_cut}}\\
\frame{\includegraphics[width=\widthplot]{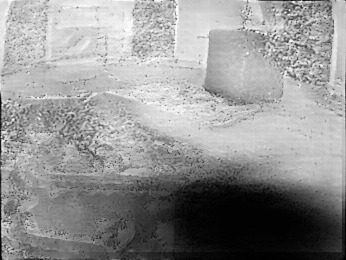}} &
\frame{\includegraphics[width=\widthplot]{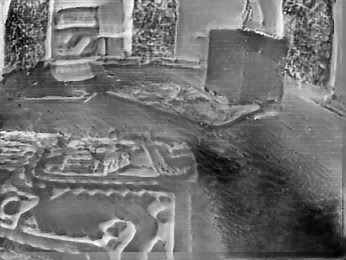}} &
\frame{\includegraphics[width=\widthplot]{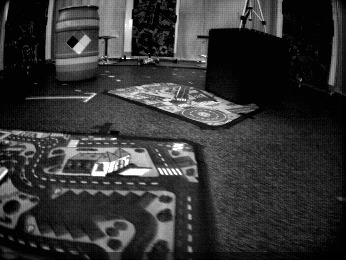}} \\

\multicolumn{3}{c}{\dsfont{outdoor\_day1\_data\_cut}}\\
\frame{\includegraphics[width=\widthplot]{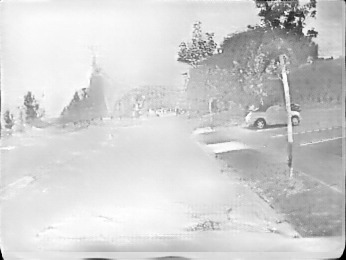}} &
\frame{\includegraphics[width=\widthplot]{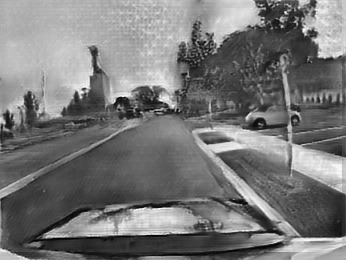}} &
\frame{\includegraphics[width=\widthplot]{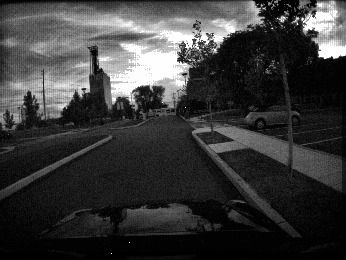}} \\

\multicolumn{3}{c}{\dsfont{outdoor\_day2\_data\_cut}}\\
\frame{\includegraphics[width=\widthplot]{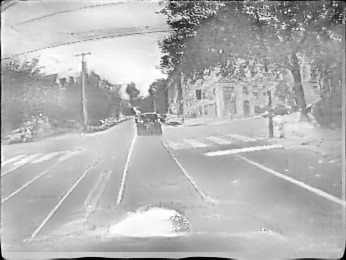}} &
\frame{\includegraphics[width=\widthplot]{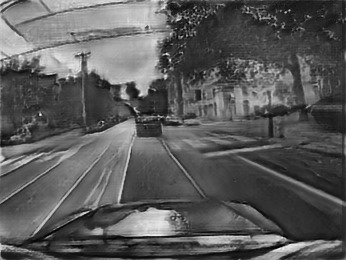}} &
\frame{\includegraphics[width=\widthplot]{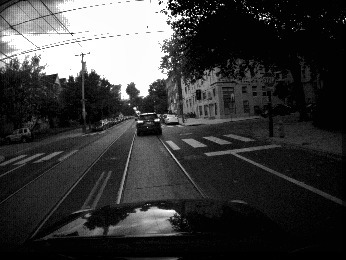}} \\

\caption{Qualitative results for MVSEC. Random selection, not cherry picked.}
\end{longtable}

\def \widthplot {0.25\linewidth}
\begin{longtable}{ccc}
	E2VID & Ours & Groundtruth\\
	\multicolumn{3}{c}{\dsfont{Fruit}}\\
	\frame{\includegraphics[width=\widthplot]{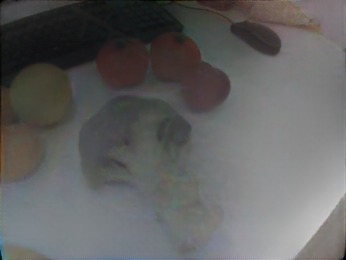}} &
	\frame{\includegraphics[width=\widthplot]{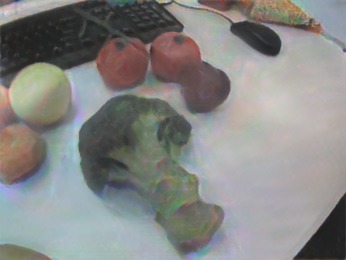}} &
	\frame{\includegraphics[width=\widthplot]{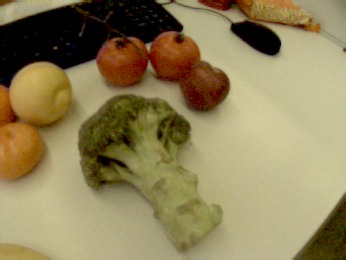}} \\
	
	\multicolumn{3}{c}{\dsfont{Keyboard}}\\
	\frame{\includegraphics[width=\widthplot]{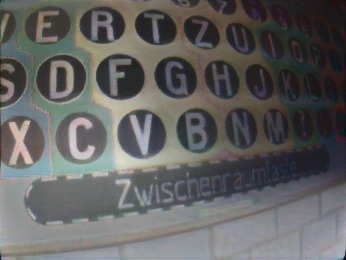}} &
	\frame{\includegraphics[width=\widthplot]{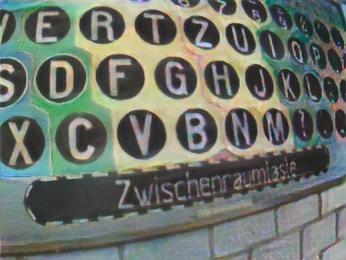}} &
	\frame{\includegraphics[width=\widthplot]{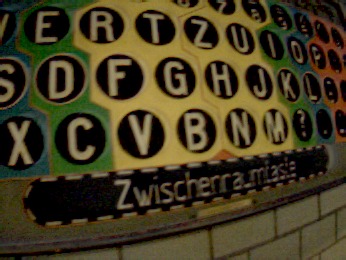}} \\
	
	\multicolumn{3}{c}{\dsfont{Carpet}}\\
	\frame{\includegraphics[width=\widthplot]{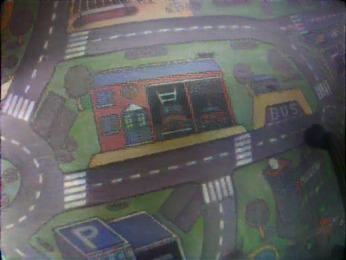}} &
	\frame{\includegraphics[width=\widthplot]{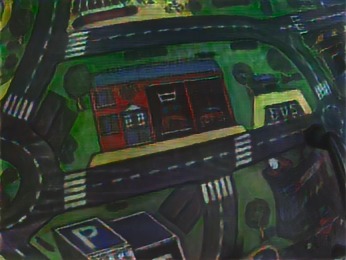}} &
	\frame{\includegraphics[width=\widthplot]{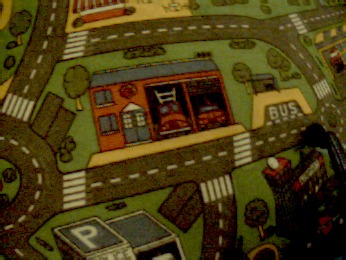}} \\
	
	\multicolumn{3}{c}{\dsfont{Jenga}}\\
	\frame{\includegraphics[width=\widthplot]{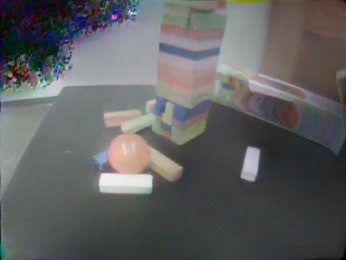}} &
	\frame{\includegraphics[width=\widthplot]{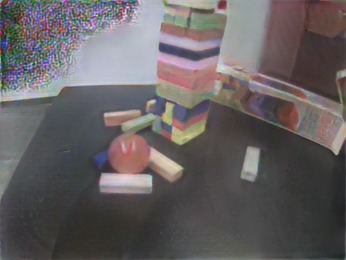}} &
	\frame{\includegraphics[width=\widthplot]{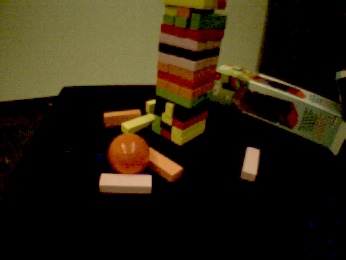}} \\
	
	\multicolumn{3}{c}{\dsfont{Object}}\\
	\frame{\includegraphics[width=\widthplot]{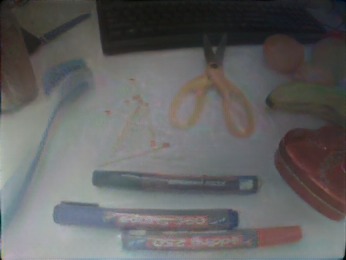}} &
	\frame{\includegraphics[width=\widthplot]{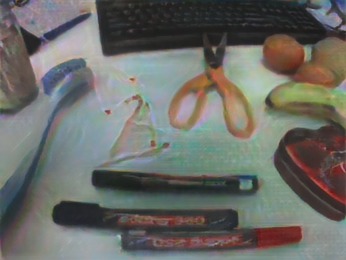}} &
	\frame{\includegraphics[width=\widthplot]{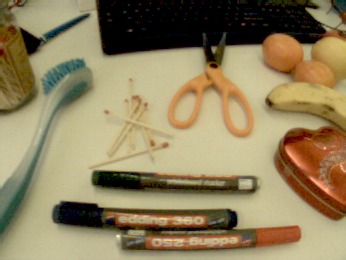}} \\
	\caption{Qualitative results for CED \cite{Scheerlinck19cvprw}. Random selection, not cherry picked.
		As a matter of interest, the Jenga sequence shows a region of the scene where there is only blank wall, so few events have been generated, resulting in the peculiar artifacts seen in the top left corner.
	}
	\label{fig:color_reconstruction}
\end{longtable}

\newpage
\def \widthplot {0.23\linewidth}
\begin{longtable}{cccc}
EVFlow & Ours & EVFlow IWE & Ours IWE \\
\multicolumn{4}{c}{\dsfont{bike\_bay\_hdr}}\\
\frame{\includegraphics[width=\widthplot]{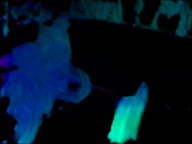}} &
\frame{\includegraphics[width=\widthplot]{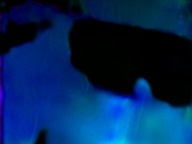}} &
\frame{\includegraphics[width=\widthplot]{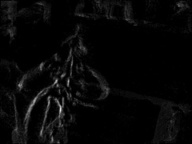}} &
\frame{\includegraphics[width=\widthplot]{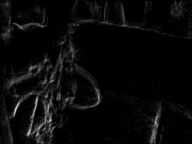}} \\

\multicolumn{4}{c}{\dsfont{boxes}}\\
\frame{\includegraphics[width=\widthplot]{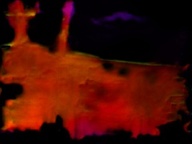}} &
\frame{\includegraphics[width=\widthplot]{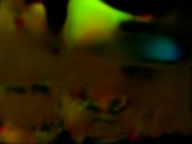}} &
\frame{\includegraphics[width=\widthplot]{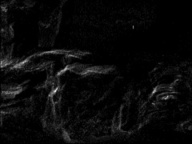}} &
\frame{\includegraphics[width=\widthplot]{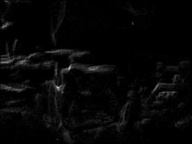}} \\

\multicolumn{4}{c}{\dsfont{desk\_6k}}\\
\frame{\includegraphics[width=\widthplot]{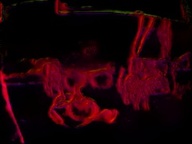}} &
\frame{\includegraphics[width=\widthplot]{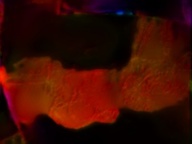}} &
\frame{\includegraphics[width=\widthplot]{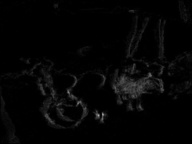}} &
\frame{\includegraphics[width=\widthplot]{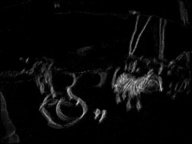}} \\

\multicolumn{4}{c}{\dsfont{desk\_fast}}\\
\frame{\includegraphics[width=\widthplot]{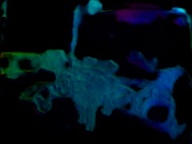}} &
\frame{\includegraphics[width=\widthplot]{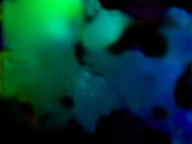}} &
\frame{\includegraphics[width=\widthplot]{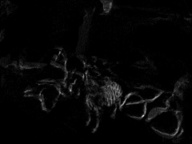}} &
\frame{\includegraphics[width=\widthplot]{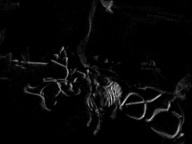}} \\

\multicolumn{4}{c}{\dsfont{desk\_hand\_only}}\\
\frame{\includegraphics[width=\widthplot]{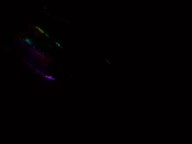}} &
\frame{\includegraphics[width=\widthplot]{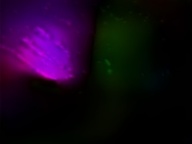}} &
\frame{\includegraphics[width=\widthplot]{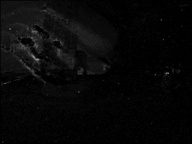}} &
\frame{\includegraphics[width=\widthplot]{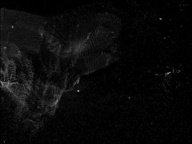}} \\

\multicolumn{4}{c}{\dsfont{desk\_slow}}\\
\frame{\includegraphics[width=\widthplot]{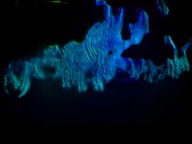}} &
\frame{\includegraphics[width=\widthplot]{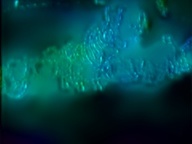}} &
\frame{\includegraphics[width=\widthplot]{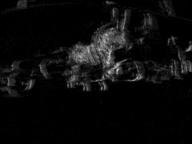}} &
\frame{\includegraphics[width=\widthplot]{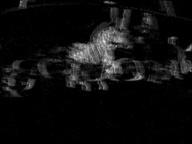}} \\

\multicolumn{4}{c}{\dsfont{engineering\_posters}}\\
\frame{\includegraphics[width=\widthplot]{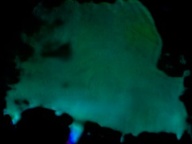}} &
\frame{\includegraphics[width=\widthplot]{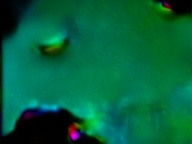}} &
\frame{\includegraphics[width=\widthplot]{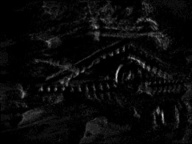}} &
\frame{\includegraphics[width=\widthplot]{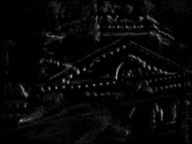}} \\

\multicolumn{4}{c}{\dsfont{high\_texture\_plants}}\\
\frame{\includegraphics[width=\widthplot]{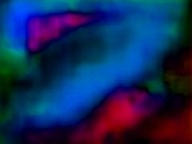}} &
\frame{\includegraphics[width=\widthplot]{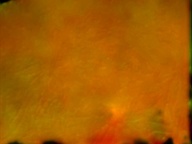}} &
\frame{\includegraphics[width=\widthplot]{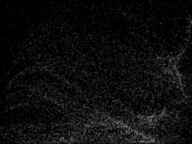}} &
\frame{\includegraphics[width=\widthplot]{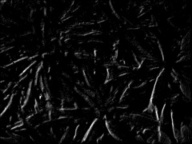}} \\

\multicolumn{4}{c}{\dsfont{poster\_pillar\_1}}\\
\frame{\includegraphics[width=\widthplot]{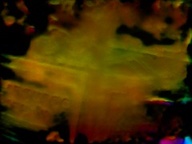}} &
\frame{\includegraphics[width=\widthplot]{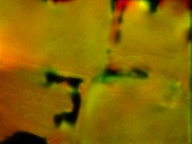}} &
\frame{\includegraphics[width=\widthplot]{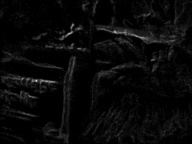}} &
\frame{\includegraphics[width=\widthplot]{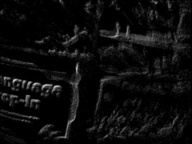}} \\

\multicolumn{4}{c}{\dsfont{poster\_pillar\_2}}\\
\frame{\includegraphics[width=\widthplot]{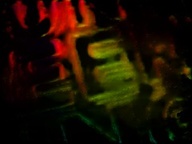}} &
\frame{\includegraphics[width=\widthplot]{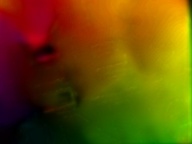}} &
\frame{\includegraphics[width=\widthplot]{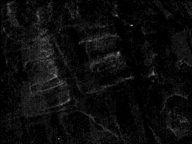}} &
\frame{\includegraphics[width=\widthplot]{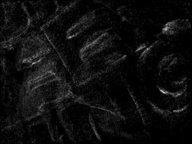}} \\

\multicolumn{4}{c}{\dsfont{reflective\_materials}}\\
\frame{\includegraphics[width=\widthplot]{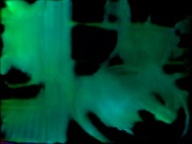}} &
\frame{\includegraphics[width=\widthplot]{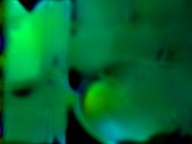}} &
\frame{\includegraphics[width=\widthplot]{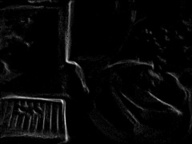}} &
\frame{\includegraphics[width=\widthplot]{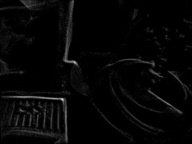}} \\

\multicolumn{4}{c}{\dsfont{slow\_and\_fast\_desk}}\\
\frame{\includegraphics[width=\widthplot]{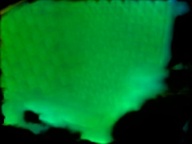}} &
\frame{\includegraphics[width=\widthplot]{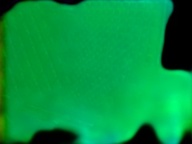}} &
\frame{\includegraphics[width=\widthplot]{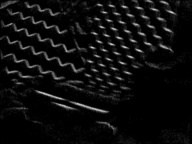}} &
\frame{\includegraphics[width=\widthplot]{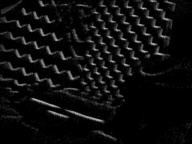}} \\

\multicolumn{4}{c}{\dsfont{slow\_hand}}\\
\frame{\includegraphics[width=\widthplot]{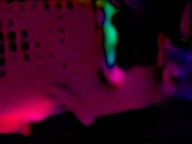}} &
\frame{\includegraphics[width=\widthplot]{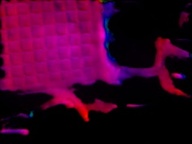}} &
\frame{\includegraphics[width=\widthplot]{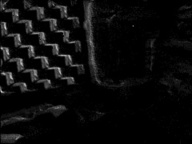}} &
\frame{\includegraphics[width=\widthplot]{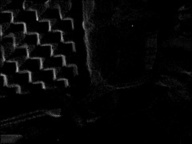}} \\

\multicolumn{4}{c}{\dsfont{still\_life}}\\
\frame{\includegraphics[width=\widthplot]{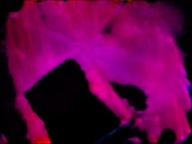}} &
\frame{\includegraphics[width=\widthplot]{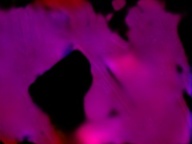}} &
\frame{\includegraphics[width=\widthplot]{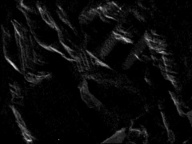}} &
\frame{\includegraphics[width=\widthplot]{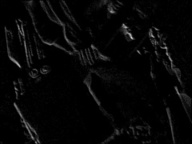}} \\

\caption{
    Qualitative results for HQFD.
	Left: optic flow vectors represented in HSV color space, right: image of warped events (IWE).
	Random selection, not cherry picked.}
\end{longtable}

\def \widthplot {0.23\linewidth}
\begin{longtable}{cccc}
EVFlow & Ours & EVFlow IWE & Ours IWE \\
\multicolumn{4}{c}{\dsfont{boxes\_6dof}}\\
\frame{\includegraphics[width=\widthplot]{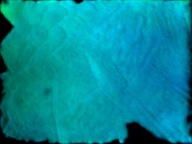}} &
\frame{\includegraphics[width=\widthplot]{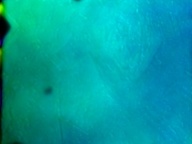}} &
\frame{\includegraphics[width=\widthplot]{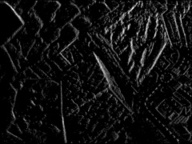}} &
\frame{\includegraphics[width=\widthplot]{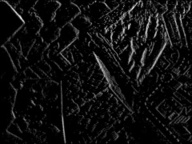}} \\

\multicolumn{4}{c}{\dsfont{calibration}}\\
\frame{\includegraphics[width=\widthplot]{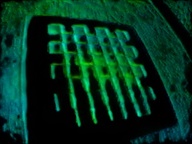}} &
\frame{\includegraphics[width=\widthplot]{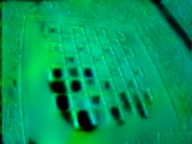}} &
\frame{\includegraphics[width=\widthplot]{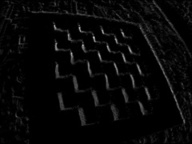}} &
\frame{\includegraphics[width=\widthplot]{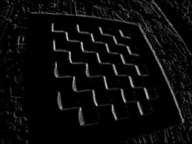}} \\

\multicolumn{4}{c}{\dsfont{dynamic\_6dof}}\\
\frame{\includegraphics[width=\widthplot]{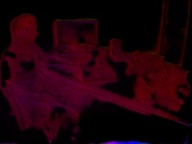}} &
\frame{\includegraphics[width=\widthplot]{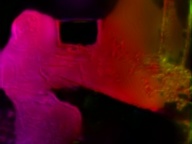}} &
\frame{\includegraphics[width=\widthplot]{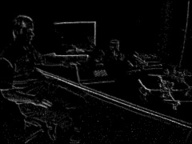}} &
\frame{\includegraphics[width=\widthplot]{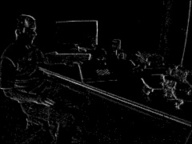}} \\

\multicolumn{4}{c}{\dsfont{office\_zigzag}}\\
\frame{\includegraphics[width=\widthplot]{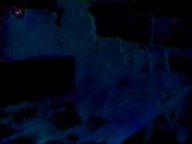}} &
\frame{\includegraphics[width=\widthplot]{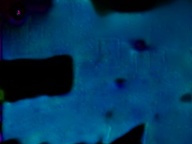}} &
\frame{\includegraphics[width=\widthplot]{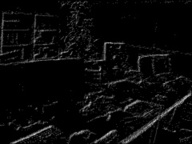}} &
\frame{\includegraphics[width=\widthplot]{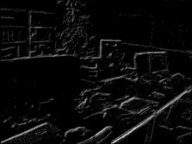}} \\

\multicolumn{4}{c}{\dsfont{poster\_6dof}}\\
\frame{\includegraphics[width=\widthplot]{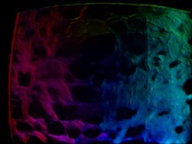}} &
\frame{\includegraphics[width=\widthplot]{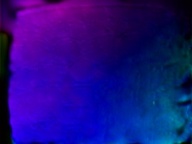}} &
\frame{\includegraphics[width=\widthplot]{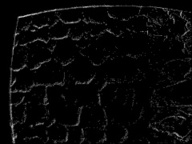}} &
\frame{\includegraphics[width=\widthplot]{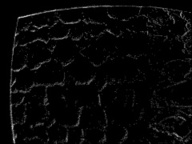}} \\

\multicolumn{4}{c}{\dsfont{shapes\_6dof}}\\
\frame{\includegraphics[width=\widthplot]{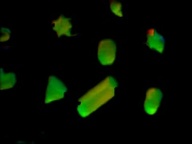}} &
\frame{\includegraphics[width=\widthplot]{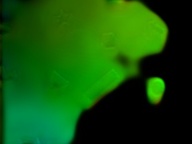}} &
\frame{\includegraphics[width=\widthplot]{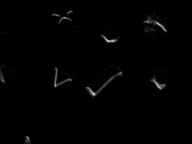}} &
\frame{\includegraphics[width=\widthplot]{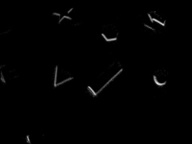}} \\

\multicolumn{4}{c}{\dsfont{slider\_depth}}\\
\frame{\includegraphics[width=\widthplot]{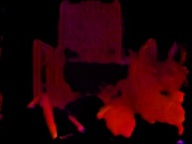}} &
\frame{\includegraphics[width=\widthplot]{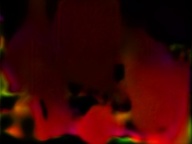}} &
\frame{\includegraphics[width=\widthplot]{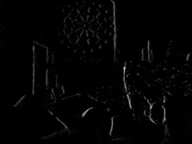}} &
\frame{\includegraphics[width=\widthplot]{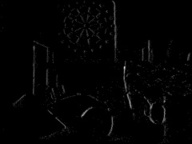}} \\

\caption{Qualitative results for IJRR. 
	Left: optic flow vectors represented in HSV color space, right: image of warped events (IWE).
	Random selection, not cherry picked.}
\end{longtable}

\def \widthplot {0.23\linewidth}
\begin{longtable}{cccc}
EVFlow & Ours & EVFlow IWE & Ours IWE \\
\multicolumn{4}{c}{\dsfont{indoor\_flying1\_data}}\\
\frame{\includegraphics[width=\widthplot]{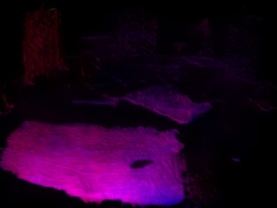}} &
\frame{\includegraphics[width=\widthplot]{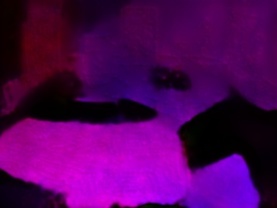}} &
\frame{\includegraphics[width=\widthplot]{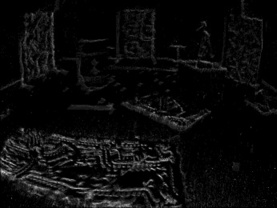}} &
\frame{\includegraphics[width=\widthplot]{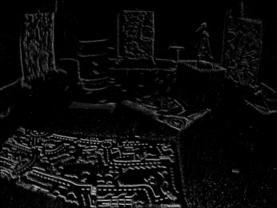}} \\

\multicolumn{4}{c}{\dsfont{indoor\_flying2\_data}}\\
\frame{\includegraphics[width=\widthplot]{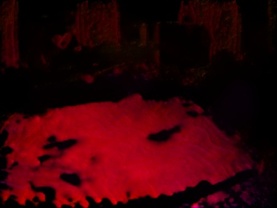}} &
\frame{\includegraphics[width=\widthplot]{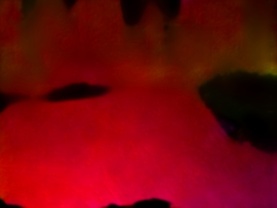}} &
\frame{\includegraphics[width=\widthplot]{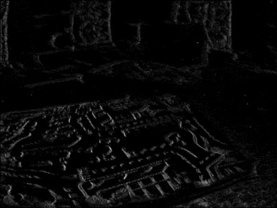}} &
\frame{\includegraphics[width=\widthplot]{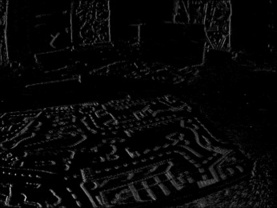}} \\

\multicolumn{4}{c}{\dsfont{indoor\_flying3\_data}}\\
\frame{\includegraphics[width=\widthplot]{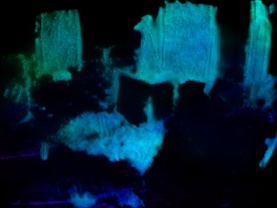}} &
\frame{\includegraphics[width=\widthplot]{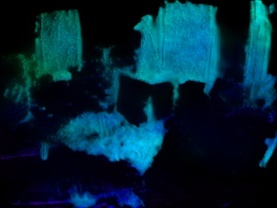}} &
\frame{\includegraphics[width=\widthplot]{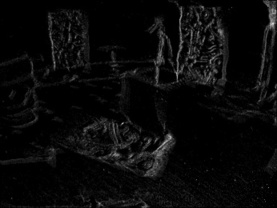}} &
\frame{\includegraphics[width=\widthplot]{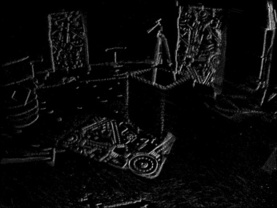}} \\

\multicolumn{4}{c}{\dsfont{indoor\_flying4\_data}}\\
\frame{\includegraphics[width=\widthplot]{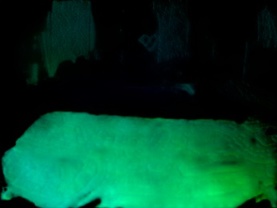}} &
\frame{\includegraphics[width=\widthplot]{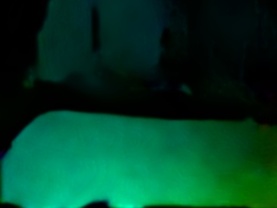}} &
\frame{\includegraphics[width=\widthplot]{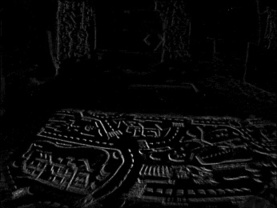}} &
\frame{\includegraphics[width=\widthplot]{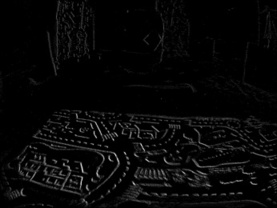}} \\

\multicolumn{4}{c}{\dsfont{outdoor\_day1\_data}}\\
\frame{\includegraphics[width=\widthplot]{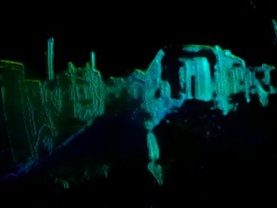}} &
\frame{\includegraphics[width=\widthplot]{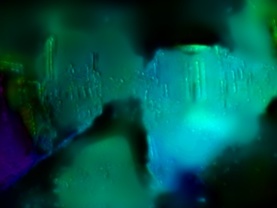}} &
\frame{\includegraphics[width=\widthplot]{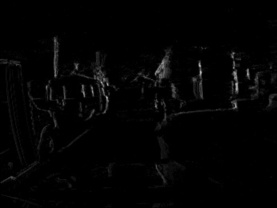}} &
\frame{\includegraphics[width=\widthplot]{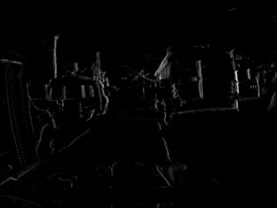}} \\

\multicolumn{4}{c}{\dsfont{outdoor\_day2\_data}}\\
\frame{\includegraphics[width=\widthplot]{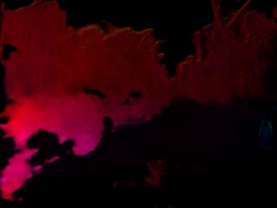}} &
\frame{\includegraphics[width=\widthplot]{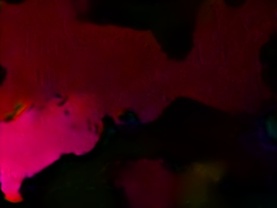}} &
\frame{\includegraphics[width=\widthplot]{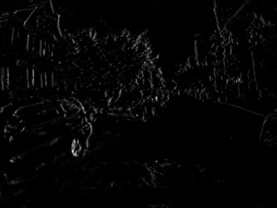}} &
\frame{\includegraphics[width=\widthplot]{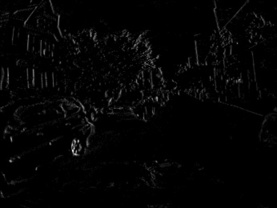}} \\

\caption{
	Qualitative results for optic flow on MVSEC.
	Left: optic flow vectors represented in HSV color space, right: image of warped events (IWE).
	Random selection, not cherry picked.}
\end{longtable}

\subsection{FireNet}

Scheerlinck \etal \cite{Scheerlinck20wacv} propose a lightweight network architecture for fast image reconstruction with an event camera (FireNet) that has \SI{99.6}{\percent} fewer parameters than E2VID \cite{Rebecq20pami} while achieving similar accuracy on IJRR \cite{Mueggler17ijrr}.
We retrain FireNet using our method and evaluate the original (FireNet) vs. retrained (FireNet+) on IJRR, MVSEC and HQF (\tablab ~\ref{tab:firenet}).
FireNet+ performs better on HQF and MVSEC though worse on IJRR.
One possible explanation is that the limited capacity of a smaller network limits generalizability over a wider distribution of data, and the original FireNet overfits to data similar to IJRR, namely low $\cth$s.
If our hypothesis is correct, it presents an additional disadvantage to small networks for event cameras.
Comprehensive evaluation (HQF + IJRR + MVSEC) reveals bigger performance gap between FireNet (\tablab \ref{tab:firenet}) and E2VID (\tablab \ref{tab:eval:all_in_one}) architectures than shown in \cite{Scheerlinck20wacv} (IJRR only).
Qualitatively (\figlab \ref{fig:firenet:hqfd}), FireNet+ looks noisier in textureless regions, while FireNet produces lower contrast images.

\begin{table}
	\caption{
		Mean MSE, SSIM \cite{Wang04tip} and LPIPS \cite{Zhang18cvprLPIPS} on our HQF dataset, IJRR \cite{Mueggler17ijrr} and MVSEC \cite{Zhu18ral}, for original FireNet vs. retrained with our method (FireNet+).}
	\newcolumntype{Z}{S[table-format=2.3,table-auto-round]}
	\centering
	\setlength{\tabcolsep}{0.6mm}
	\renewcommand*{\arraystretch}{1.1}
	\newcommand{\smallspace}{2mm}
	\small
	\resizebox{\textwidth}{!}{%
	\begin{tabular}{@{}lcZZZcZZZcZZZcr@{}}
		\toprule
		\multirow{2}[3]{*}{Model} && \multicolumn{3}{c}{HQF} && \multicolumn{3}{c}{IJRR} && \multicolumn{3}{c}{MVSEC} \\
		\cmidrule(l{\smallspace}r{\smallspace}){3-5} \cmidrule(l{\smallspace}r{\smallspace}){7-9} \cmidrule(l{\smallspace}r{\smallspace}){11-13}
		&& {MSE} & {SSIM} & {LPIPS} && {MSE} & {SSIM} & {LPIPS} && {MSE} & {SSIM} & {LPIPS} \\
		\midrule
		FireNet && 0.0521 & \bfseries 0.5136 & 0.3865 && \bfseries 0.0554 & \bfseries 0.6296 & \bfseries 0.2566 && 0.1820 & \bfseries 0.3201 & 0.5944 \\ 
		FireNet+ && \bfseries 0.0485 & 0.4774 & \bfseries 0.3494 && 0.0579 & 0.5034 & 0.3270 && \bfseries 0.1568 & 0.2884 & \bfseries 0.5507 \\  
		\bottomrule	
	\end{tabular}
}
	\label{tab:firenet}
\end{table}

\def \widthplot {0.25\linewidth}
\begin{longtable}{ccc}
FireNet & FireNet+ & Groundtruth\\
\multicolumn{3}{c}{\dsfont{bike\_bay\_hdr}}\\
\frame{\includegraphics[width=\widthplot]{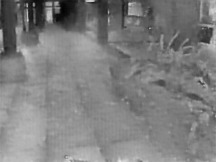}} &
\frame{\includegraphics[width=\widthplot]{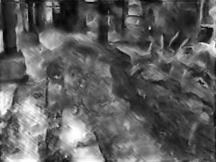}} &
\frame{\includegraphics[width=\widthplot]{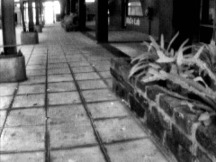}} \\

\multicolumn{3}{c}{\dsfont{boxes}}\\
\frame{\includegraphics[width=\widthplot]{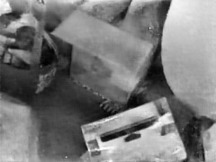}} &
\frame{\includegraphics[width=\widthplot]{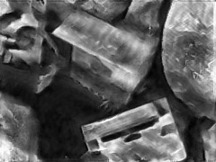}} &
\frame{\includegraphics[width=\widthplot]{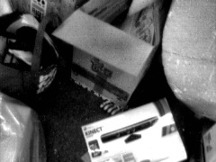}} \\

\multicolumn{3}{c}{\dsfont{desk\_6k}}\\
\frame{\includegraphics[width=\widthplot]{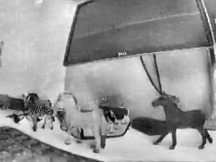}} &
\frame{\includegraphics[width=\widthplot]{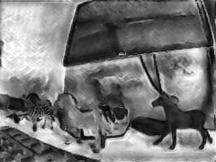}} &
\frame{\includegraphics[width=\widthplot]{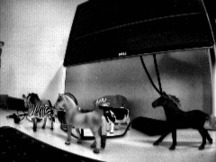}} \\

\multicolumn{3}{c}{\dsfont{desk\_fast}}\\
\frame{\includegraphics[width=\widthplot]{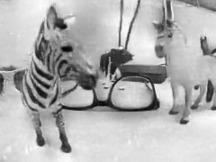}} &
\frame{\includegraphics[width=\widthplot]{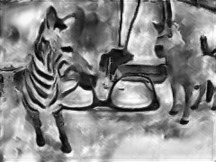}} &
\frame{\includegraphics[width=\widthplot]{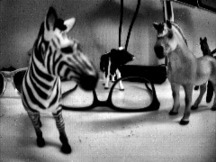}} \\

\multicolumn{3}{c}{\dsfont{desk\_hand\_only}}\\
\frame{\includegraphics[width=\widthplot]{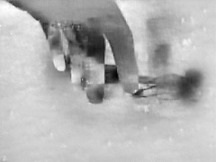}} &
\frame{\includegraphics[width=\widthplot]{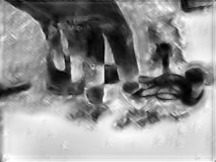}} &
\frame{\includegraphics[width=\widthplot]{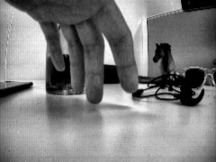}} \\

\multicolumn{3}{c}{\dsfont{desk\_slow}}\\
\frame{\includegraphics[width=\widthplot]{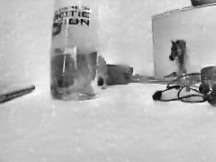}} &
\frame{\includegraphics[width=\widthplot]{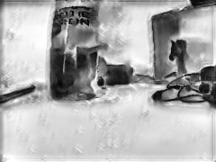}} &
\frame{\includegraphics[width=\widthplot]{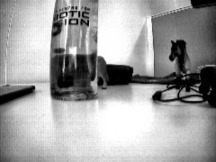}} \\

\multicolumn{3}{c}{\dsfont{engineering\_posters}}\\
\frame{\includegraphics[width=\widthplot]{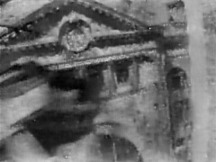}} &
\frame{\includegraphics[width=\widthplot]{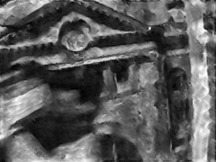}} &
\frame{\includegraphics[width=\widthplot]{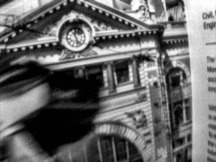}} \\

\multicolumn{3}{c}{\dsfont{high\_texture\_plants}}\\
\frame{\includegraphics[width=\widthplot]{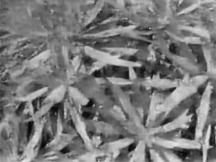}} &
\frame{\includegraphics[width=\widthplot]{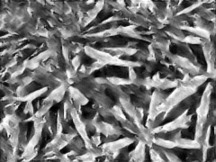}} &
\frame{\includegraphics[width=\widthplot]{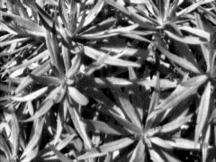}} \\

\multicolumn{3}{c}{\dsfont{poster\_pillar\_1}}\\
\frame{\includegraphics[width=\widthplot]{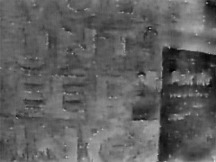}} &
\frame{\includegraphics[width=\widthplot]{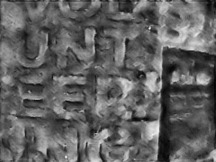}} &
\frame{\includegraphics[width=\widthplot]{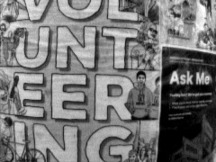}} \\

\multicolumn{3}{c}{\dsfont{poster\_pillar\_2}}\\
\frame{\includegraphics[width=\widthplot]{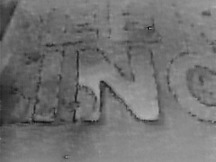}} &
\frame{\includegraphics[width=\widthplot]{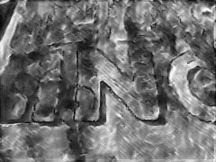}} &
\frame{\includegraphics[width=\widthplot]{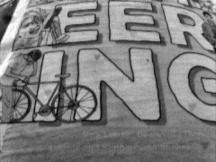}} \\

\multicolumn{3}{c}{\dsfont{reflective\_materials}}\\
\frame{\includegraphics[width=\widthplot]{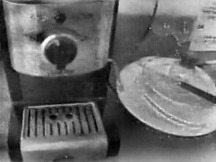}} &
\frame{\includegraphics[width=\widthplot]{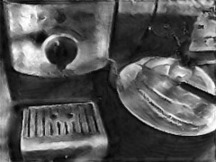}} &
\frame{\includegraphics[width=\widthplot]{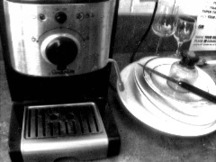}} \\

\multicolumn{3}{c}{\dsfont{slow\_and\_fast\_desk}}\\
\frame{\includegraphics[width=\widthplot]{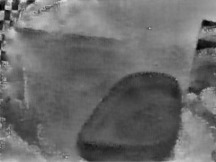}} &
\frame{\includegraphics[width=\widthplot]{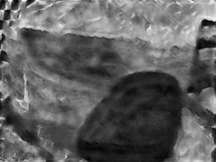}} &
\frame{\includegraphics[width=\widthplot]{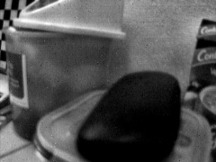}} \\

\multicolumn{3}{c}{\dsfont{slow\_hand}}\\
\frame{\includegraphics[width=\widthplot]{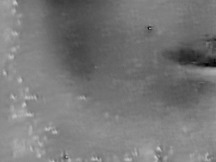}} &
\frame{\includegraphics[width=\widthplot]{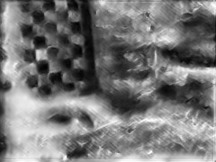}} &
\frame{\includegraphics[width=\widthplot]{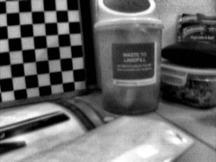}} \\

\multicolumn{3}{c}{\dsfont{still\_life}}\\
\frame{\includegraphics[width=\widthplot]{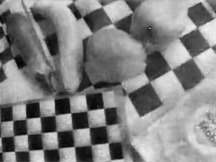}} &
\frame{\includegraphics[width=\widthplot]{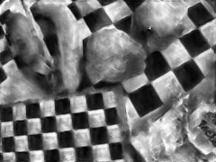}} &
\frame{\includegraphics[width=\widthplot]{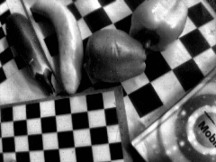}} \\

\caption{Qualitative results for HQFD. Random selection, not cherry picked.}
\label{fig:firenet:hqfd}
\end{longtable}

\def \widthplot {0.25\linewidth}
\begin{longtable}{ccc}
FireNet & FireNet+ & Groundtruth\\
\multicolumn{3}{c}{\dsfont{boxes\_6dof\_cut}}\\
\frame{\includegraphics[width=\widthplot]{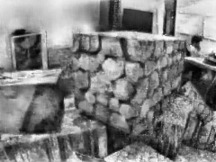}} &
\frame{\includegraphics[width=\widthplot]{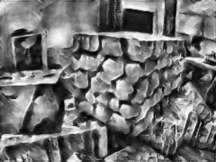}} &
\frame{\includegraphics[width=\widthplot]{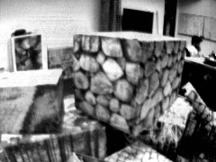}} \\

\multicolumn{3}{c}{\dsfont{calibration\_cut}}\\
\frame{\includegraphics[width=\widthplot]{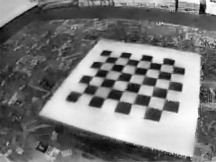}} &
\frame{\includegraphics[width=\widthplot]{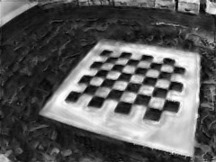}} &
\frame{\includegraphics[width=\widthplot]{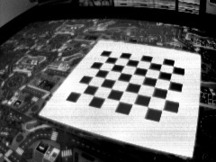}} \\

\multicolumn{3}{c}{\dsfont{dynamic\_6dof\_cut}}\\
\frame{\includegraphics[width=\widthplot]{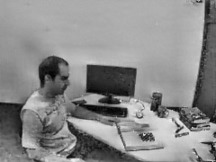}} &
\frame{\includegraphics[width=\widthplot]{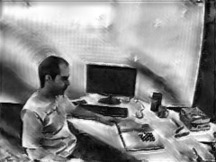}} &
\frame{\includegraphics[width=\widthplot]{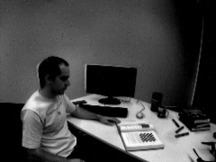}} \\

\multicolumn{3}{c}{\dsfont{office\_zigzag\_cut}}\\
\frame{\includegraphics[width=\widthplot]{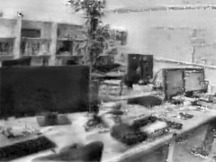}} &
\frame{\includegraphics[width=\widthplot]{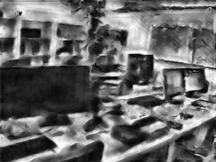}} &
\frame{\includegraphics[width=\widthplot]{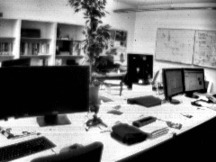}} \\

\multicolumn{3}{c}{\dsfont{poster\_6dof\_cut}}\\
\frame{\includegraphics[width=\widthplot]{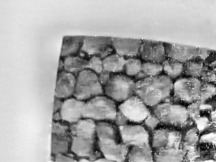}} &
\frame{\includegraphics[width=\widthplot]{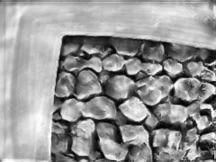}} &
\frame{\includegraphics[width=\widthplot]{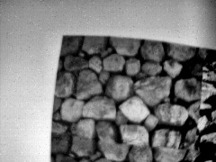}} \\

\multicolumn{3}{c}{\dsfont{shapes\_6dof\_cut}}\\
\frame{\includegraphics[width=\widthplot]{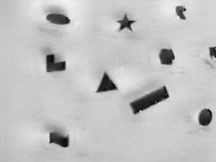}} &
\frame{\includegraphics[width=\widthplot]{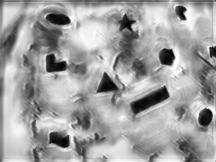}} &
\frame{\includegraphics[width=\widthplot]{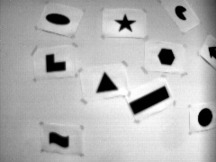}} \\

\multicolumn{3}{c}{\dsfont{slider\_depth\_cut}}\\
\frame{\includegraphics[width=\widthplot]{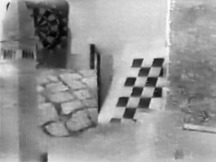}} &
\frame{\includegraphics[width=\widthplot]{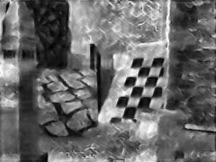}} &
\frame{\includegraphics[width=\widthplot]{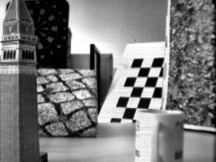}} \\

\caption{Qualitative results for IJRR. Random selection, not cherry picked.}
\label{fig:firenet:ijrr}
\end{longtable}

\def \widthplot {0.25\linewidth}
\begin{longtable}{ccc}
FireNet & FireNet+ & Groundtruth\\
\multicolumn{3}{c}{\dsfont{indoor\_flying1\_data\_cut}}\\
\frame{\includegraphics[width=\widthplot]{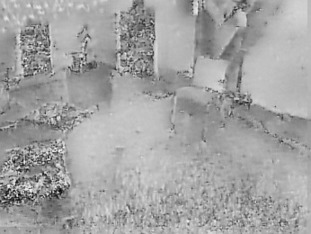}} &
\frame{\includegraphics[width=\widthplot]{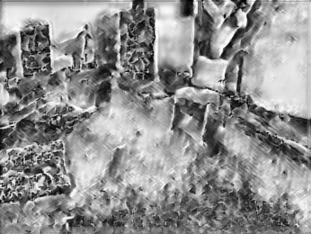}} &
\frame{\includegraphics[width=\widthplot]{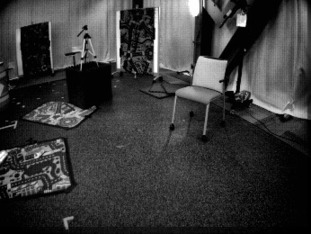}} \\

\multicolumn{3}{c}{\dsfont{indoor\_flying2\_data\_cut}}\\
\frame{\includegraphics[width=\widthplot]{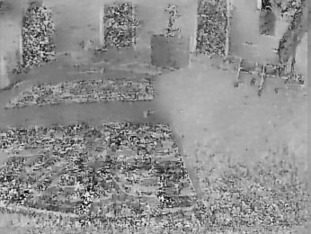}} &
\frame{\includegraphics[width=\widthplot]{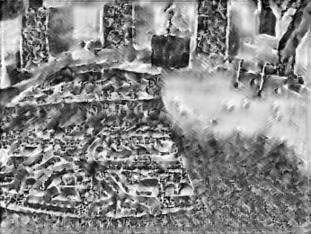}} &
\frame{\includegraphics[width=\widthplot]{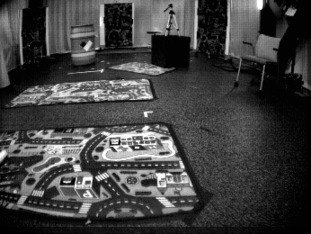}} \\

\multicolumn{3}{c}{\dsfont{indoor\_flying3\_data\_cut}}\\
\frame{\includegraphics[width=\widthplot]{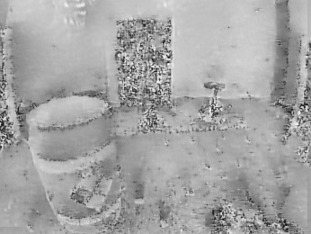}} &
\frame{\includegraphics[width=\widthplot]{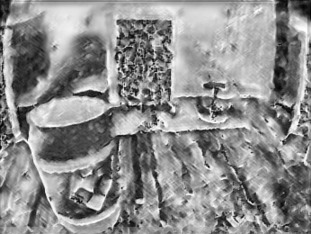}} &
\frame{\includegraphics[width=\widthplot]{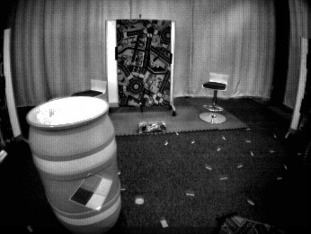}} \\

\multicolumn{3}{c}{\dsfont{indoor\_flying4\_data\_cut}}\\
\frame{\includegraphics[width=\widthplot]{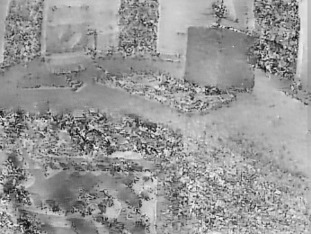}} &
\frame{\includegraphics[width=\widthplot]{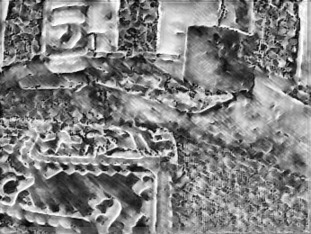}} &
\frame{\includegraphics[width=\widthplot]{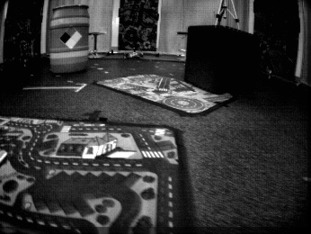}} \\

\multicolumn{3}{c}{\dsfont{outdoor\_day1\_data\_cut}}\\
\frame{\includegraphics[width=\widthplot]{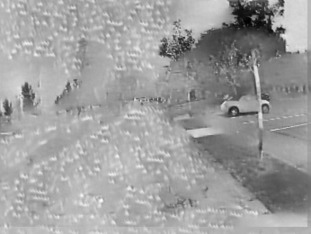}} &
\frame{\includegraphics[width=\widthplot]{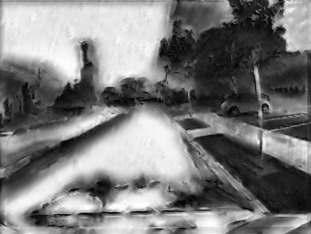}} &
\frame{\includegraphics[width=\widthplot]{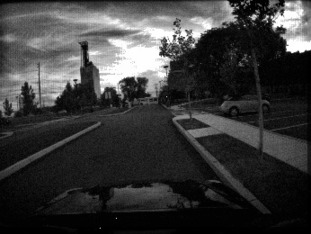}} \\

\multicolumn{3}{c}{\dsfont{outdoor\_day2\_data\_cut}}\\
\frame{\includegraphics[width=\widthplot]{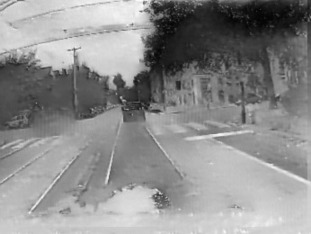}} &
\frame{\includegraphics[width=\widthplot]{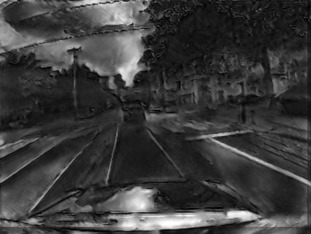}} &
\frame{\includegraphics[width=\widthplot]{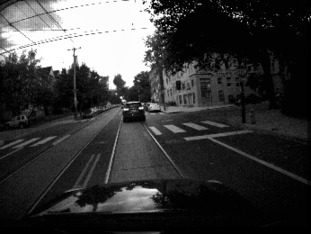}} \\

\caption{Qualitative results for MVSEC. Random selection, not cherry picked.}
\label{fig:firenet:mvsec}
\end{longtable}

\end{document}